\documentclass[sn-apa]{sn-jnl}

\jyear{2023}%

\theoremstyle{thmstyleone}%
\theoremstyle{thmstyletwo}%

\theoremstyle{thmstylethree}%

\usepackage{amsmath,amsfonts,amssymb,amsthm,epsfig,url,epstopdf,array, xparse, bm}
\usepackage{graphicx}
\usepackage{url}

\usepackage{booktabs}
\usepackage{mathtools}

\usepackage{footnote}
\usepackage{tabularx}
\usepackage{multirow, makecell}
\usepackage{textcomp}
\usepackage{subcaption}
\usepackage{patchcmd}
\usepackage{etoolbox}
\usepackage{optidef}
\usepackage{color}
\usepackage{breqn}

\begin{document}

\title[Constrained multi-objective optimization of process design parameters...]{Constrained multi-objective optimization of process design parameters in settings with scarce data: an application to adhesive bonding}


\author*[1,2]{\fnm{Alejandro} \sur{Morales-Hernández}}\email{alejandro.moraleshernandez@uhasselt.be}

\author[1,2,3]{\fnm{Sebastian} \sur{Rojas Gonzalez}}\email{sebastian.rojasgonzalez@ugent.be}

\author[1,2]{\fnm{Inneke} \sur{Van Nieuwenhuyse}}\email{inneke.vannieuwenhuyse@uhasselt.be}

\author[3]{\fnm{Ivo} \sur{Couckuyt}}\email{ivo.couckuyt@ugent.be}

\author[4]{\fnm{Jeroen} \sur{Jordens}}\email{jeroen.jordens@flandersmake.be}

\author[4]{\fnm{Maarten} \sur{Witters}}\email{maarten.witters@flandersmake.be}

\author[4]{\fnm{Bart} \sur{Van Doninck}}\email{bart.vandoninck@flandersmake.be}

\affil*[1]{\orgdiv{Flanders Make@UHasselt and Data Science Institute}, \orgname{Hasselt University}, \orgaddress{\street{Agoralaan}, \city{Diepenbeek}, \postcode{3590}, \country{Belgium}}}


\affil[2]{\orgdiv{Computational Mathematics}, \orgname{Faculty of Sciences, Hasselt University}, \orgaddress{\street{Agoralaan}, \city{Diepenbeek}, \postcode{3590}, \country{Belgium}}}

\affil[3]{\orgdiv{Surrogate Modeling Lab}, \orgname{Ghent University-IMEC}, \orgaddress{\street{Technologiepark-Zwijnaarde}, \city{Ghent}, \postcode{9052}, \country{Belgium}}}

\affil[4]{\orgdiv{CoDesignS}, \orgname{Flanders Make}, \orgaddress{\street{Oude Diestersebaan}, \city{Lommel}, \postcode{3920}, \country{Belgium}}}


\abstract{Adhesive joints are increasingly used in industry for a wide variety of applications because of their favorable characteristics such as high strength-to-weight ratio, design flexibility, limited stress concentrations, planar force transfer, good damage tolerance, and fatigue resistance. Finding the optimal process parameters for an adhesive bonding process is challenging: the optimization is inherently multi-objective (aiming to maximize break strength while minimizing cost), constrained (the process should not result in any visual damage to the materials, and stress tests should not result in failures that are adhesion-related), and uncertain (testing the same process parameters several times may lead to different break strengths). Real-life physical experiments in the lab are expensive to perform. Traditional evolutionary approaches (such as genetic algorithms) are then ill-suited to solve the problem, due to the prohibitive amount of experiments required for evaluation. Although Bayesian optimization-based algorithms are preferred to solve such expensive problems, few methods consider the optimization of more than one (noisy) objective and several constraints at the same time. In this research, we successfully applied specific machine learning techniques (Gaussian Process Regression) to emulate the objective and constraint functions based on a \emph{limited} amount of experimental data. The techniques are embedded in a Bayesian optimization algorithm, which succeeds in detecting Pareto-optimal process settings in a highly efficient way (i.e., requiring a limited number of physical experiments). }




\maketitle

\section{Introduction}
\label{sec:introduction}

Adhesive bonding is the engineering process of joining two surfaces together by a non-metallic substance \citep{brockmann2008adhesive}. This process occurs frequently in many engineering design contexts, such as the automotive industry \citep{met2020}, electronics \citep{licari2011adhesives}, and aeronautics \citep{correia2018effect}. It is a complex process, in which several physical and chemical processes occur simultaneously \citep{daSilva2011,Pocius2021}, with outcomes that are influenced by many factors, such as environmental conditions, material specifications, and specific process settings. Process optimization is therefore traditionally performed by experts, based on acquired knowledge and extensive experimental campaigns \citep{Budhe2017,Habenicht2008}. Yet, even for experts, optimizing adhesive bonding parameters is a difficult task, due to the complex (and, essentially, black box) relationships between the process parameters, and the case-specific nature of the bonding process. As a consequence, real physical experiments are required to detect the optimal settings for each specific case. These tend to be costly in terms of time (time to prepare the parts to be joined, time to prepare the adhesive, to join both materials, and to allow the adhesive to gain its final strength), and in terms of manual labor. Moreover, data from one industrial bonding process cannot be used to optimize another process, as not only materials and adhesives may differ, but also production process specifications. As a result, optimization of an adhesive bonding process is challenging and costly, even with expert knowledge available. Moreover, the experimental approach may easily yield suboptimal results with respect to other relevant performance metrics, such as production costs. Thus, the problem is also inherently \emph{multi-objective}. 

As observed by \cite{chiarello2021data}, evolutionary approaches (in particular, genetic algorithms) are currently the most common data science technique in engineering design optimization. This also holds for multi-objective engineering design problems (a.o., bonding problems). Evolutionary multi-objective algorithms (EMOAs) are applicable to black-box optimization problems and have proven to be effective derivative-free optimizers \citep{Zhou11}. The \textit{Non-dominated Sorting Genetic Algorithm (NSGA-II)} proposed by \cite{deb2002fast}, stands out as one of the most widely used algorithms in industrial process cases. For instance, \cite{corbett2017numerical} used NSGA-II in combination with a Gaussian Process surrogate model to simultaneously optimize two objectives (force sustained by the joint and area beneath the load-displacement curve) for a novel concept for joining materials. \cite{labbe2012multi} studied designs for optimal bonding of TSL (tubular single-lap) joints using NSGA-II, highlighting the conflicting nature of mass-related and stress-related objectives. 

Relative to the unconstrained case, the literature on algorithms for solving \emph{constrained} multi-objective problems is scarce. Some of the well-known algorithms are the Constrained Non-dominated Sorting Genetic Algorithm II (C-NSGA-II, \citealt{deb2002fast,brownlee2015constrained}), the Constrained Multi-Objective Evolutionary Algorithm based on Decomposition (C-MOEA/D, \citealt{jain2013evolutionary}), the Two-Archive Evolutionary Algorithm for Constrained multi-objective optimization (C-TAEA, \citealt{li2018two}), and the Constrained Particle Swarm Optimization algorithm for Multi-Objective problems (C-MOPSO, \citealt{gao2012constrained}). These algorithms differ in the way in which they handle constraints: while the first three algorithms handle the constraints during the non-dominated sort, the latter adopts a penalization function approach. We refer the reader to the original publications for further details about the implementation of these algorithms.

EMOAs are indeed an effective tool for black-box optimization, especially if the optimization problem is difficult to model; however, they are \emph{data-inefficient} (i.e., they require numerous function evaluations), which makes them ill-suited for optimizing design problems that require (computational or experimental) \emph{expensive} data. Even when an emulator or surrogate model is used to mitigate this issue (see e.g., \cite{chugh2017survey}), the search process in this type of algorithm remains largely based on random alterations of (promising) solutions, with convergence speeds that are sensitive to the choice of user-defined (and often problem-specific) parameters. Moreover, the measurement of the objectives is usually affected by noise and this is often neglected during the optimization, leading to suboptimal solutions.

For the sub-class of low-dimensional black-box problems that are expensive to evaluate, Bayesian optimization (BO) has emerged as a powerful alternative, with applications ranging from hyperparameter tuning of deep learning models, to design optimization in engineering, and stochastic optimization in operational research (see \cite{frazier2018bayesian} for a comprehensive review). Several Bayesian Multiobjective Optimization (BMO) methods have been proposed to solve complex decision-making problems involving multiple objectives that are expensive to evaluate \citep{Seb19rev}. Only a few methods have considered the constrained case (see e.g., \citealt{hernandez2015predictive,gelbart2016general,garrido2019predictive}), and even less literature has considered noisy objectives and/or constraints (see e.g., \citealt{feliot2017bayesian,Seb19}). This paper illustrates the power of BMO approaches for the (noisy and constrained) optimization of a novel adhesive bonding process. The approach builds on the recent work of \cite{Seb19}; the main contributions include:

\begin{itemize}
\item The formulation of an acquisition function that combines the expected improvement over the objectives and the constraint feasibility, where both the objective and constraints are black-box, noisy and expensive to evaluate. This provides an explainable approach, as opposed to the upfront parameter choices that guide the generation of successive populations in evolutionary algorithms. 

\item The use of a Gaussian Process (GP) surrogate that explicitly accounts for the noise that is typically present in the outcomes of the experiments. Here, outcomes cannot be observed with perfect accuracy, and the magnitude of the noise may vary between different configurations; thus, we have \emph{input-dependent} (i.e., heterogeneous) noise. This issue is commonly overlooked in engineering optimization problems, by simply assuming that noise is non-existent (or, at best, homogeneous). 

\item A full BMO algorithm that is able to obtain \emph{better} configurations than state-of-the-art EMOAs and surrogate-assisted EMOAs. The approach is particularly meant for settings where the analyst can only afford a very limited number of observations (as is the case with costly physical experiments in a lab). As it is general, it may constitute a useful tool also in other data-constrained engineering design settings.

\end{itemize}

The remainder of this article is organized as follows. Section \ref{sec:moo} discusses the main concepts in multi-objective optimization, followed by a description of the bonding process problem under study. Section \ref{sec:moo-meth} details the relevant background in Bayesian optimization, and the proposed algorithms. Section \ref{sec:experiment_design} discusses the design of experiments, while Section \ref{sec:results} discusses the results. Finally, Section \ref{sec:conclusions} summarizes the findings and highlights some future research directions. 

\section{Multi-objective optimization: main concepts and problem description}
\label{sec:moo}

\subsection{Multi-objective optimization: main concepts }
\label{sec:moo_concepts}
 In general, a multi-objective optimization problem can be defined as (assuming minimization of all the objectives):

\begin{mini}|s|
{}{\left[ f_1(\mathbf{x}), \dots, f_m(\mathbf{x}) \right ]}{\label{eq:moo}}{}
\addConstraint{g_i(\mathbf{x}) \leq 0, i=1,2, \dots, p}
\end{mini}

\noindent where $\mathbf{x}=[x_1, \dots, x_d]^T$ is a vector of $d$ decision variables in the decision space $D$, $f: D \rightarrow \mathbb{R}^m$ is a vector-valued function yielding the $m$ objectives, and the function $g_i$ ($i=1,...,p$) defines inequality constraints. The goal of a multi-objective problem is to find the set of \textit{non-dominated} or \textit{Pareto-optimal} solutions. A solution $\mathbf{x}^{(1)}$ is said to dominate another solution $\mathbf{x}^{(2)}$ when it performs better than the latter on at least one objective, while not performing worse on any of the other objectives \citep{Miet99}. More formally, for $\mathbf{x}^{(1)}$ and $\mathbf{x}^{(2)}$ two vectors in $D$:

\begin{itemize}
    \item $\mathbf{x}^{(1)} \prec \mathbf{x}^{(2)}$ means $\mathbf{x}^{(1)}$ dominates $\mathbf{x}^{(2)}$ iff $f_j(\mathbf{x}^{(1)}) \leq f_j(\mathbf{x}^{(2)}), \forall j \in \{1, \dots, m\}$, and $\exists  j \in \{1, \dots, m\}$ such that $f_j(\mathbf{x}^{(1)}) < f_j(\mathbf{x}^{(2)})$
    
    \item  $\mathbf{x}^{(1)} \prec \prec \mathbf{x}^{(2)}$ means $\mathbf{x}^{(1)}$ strictly dominates $\mathbf{x}^{(2)}$ iff $f_j(\mathbf{x}^{(1)}) < f_j(\mathbf{x}^{(2)}), \forall j \in \{1, \dots, m\}$
\end{itemize}

\begin{figure}[!htbp]
\centering
\begin{subfigure}[b]{0.45\textwidth}
\centering
\includegraphics[width=5cm]{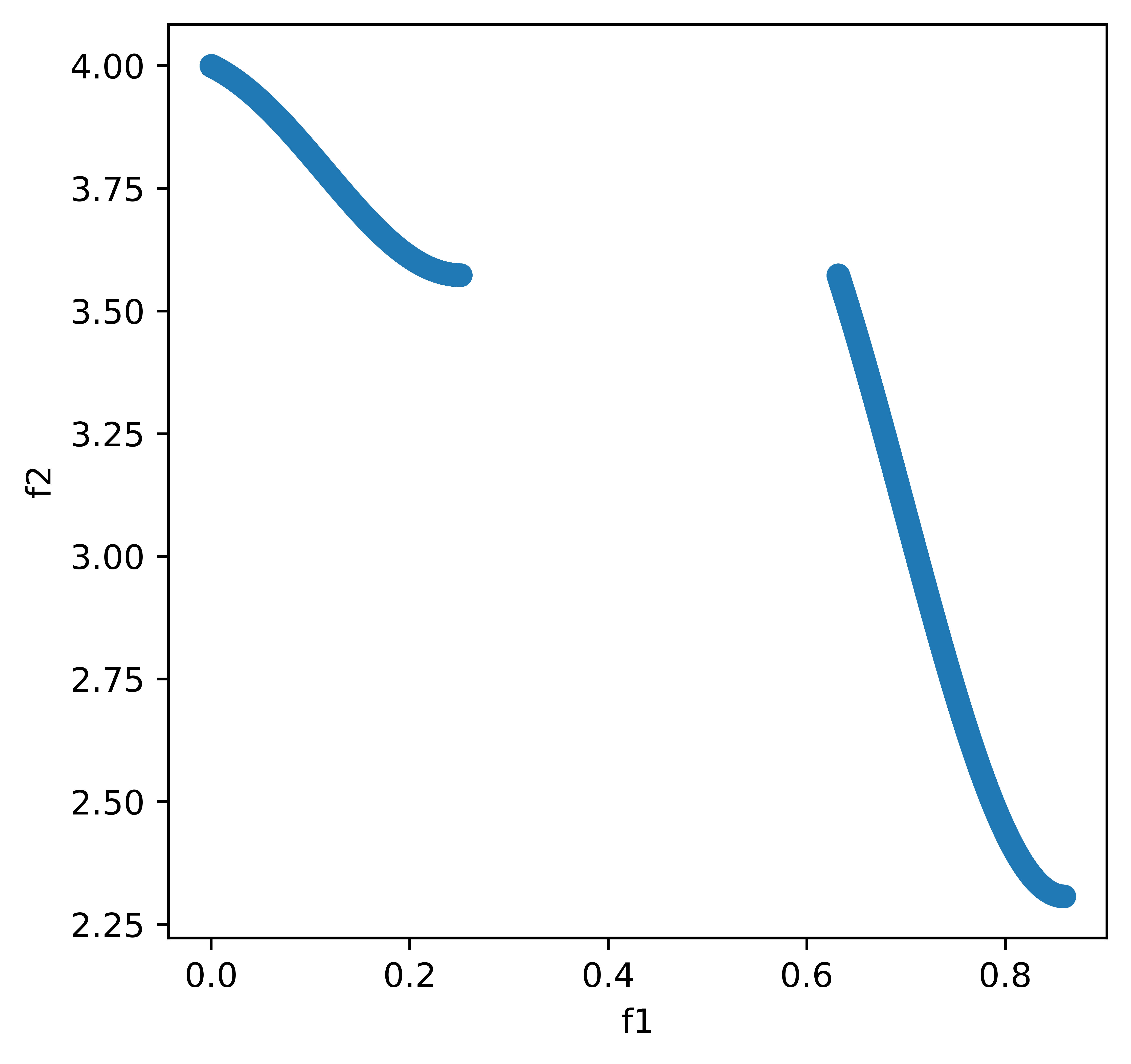} 
\caption{}
\label{fig:2d_function}
\end{subfigure}
\hfill
\begin{subfigure}[b]{0.45\textwidth}
\centering
\includegraphics[width=5.5cm, height=5cm]{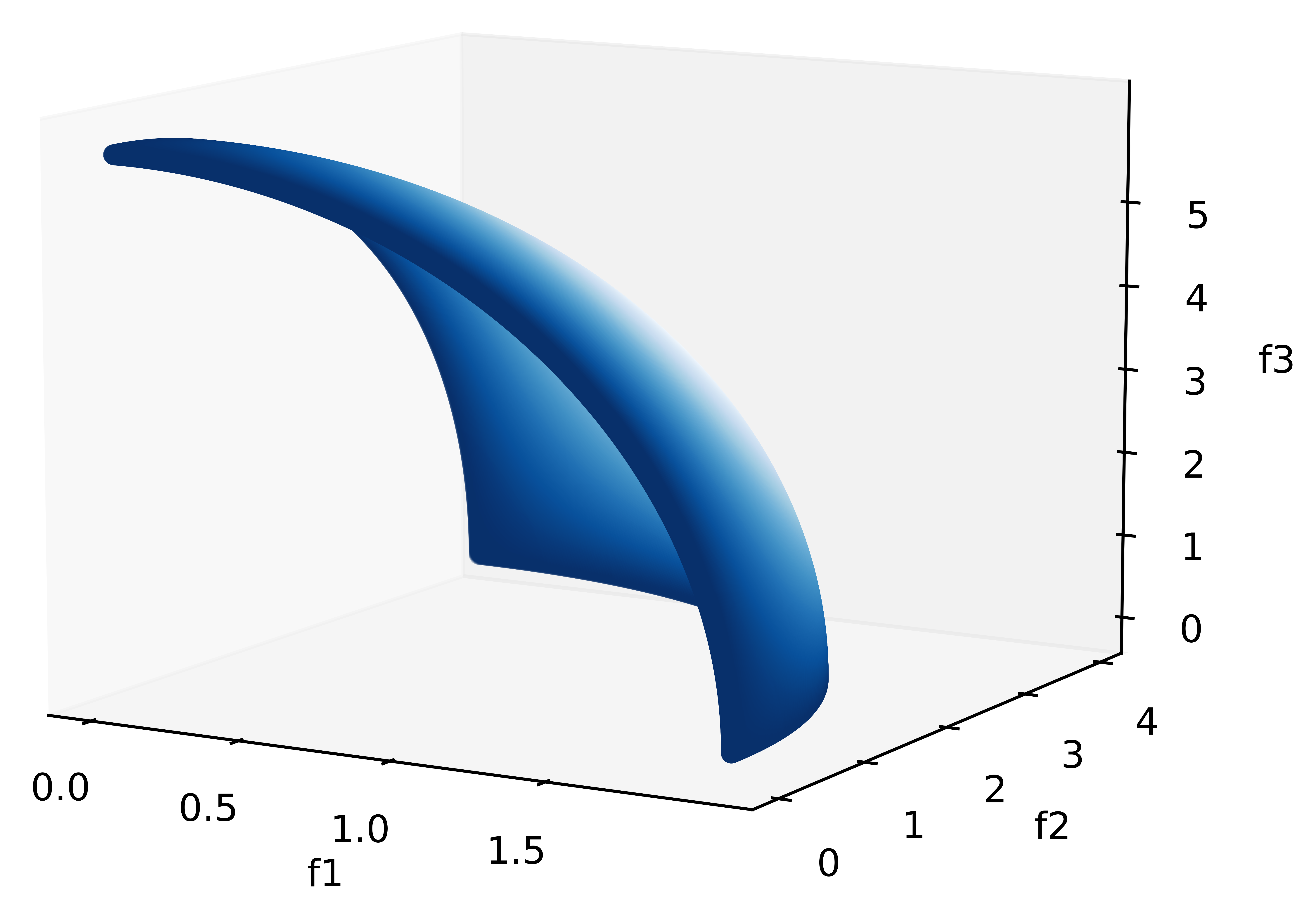} 
\caption{}
\label{fig:3d_function}
\end{subfigure}
\caption{Illustration of (a) a 2-objective problem with a disconnected Pareto front, (b) a 3-objective problem with a concave Pareto front}
\label{fig:pf}
\end{figure}

The optimal solutions for the individual objectives are  usually not of interest to the decision-maker, since these only reflect the extremes of the so-called \emph{Pareto front} (i.e., the evaluation of the Pareto set in the objective space). The geometry of the Pareto front can be very diverse (see, e.g., Figure \ref{fig:pf} for two examples generated from artificial test problems), and is usually unknown upfront. Often, it contains infinitely many solutions; the multi-objective optimization algorithm then tries to approximate this front with a finite discrete set. The solution to be implemented in practice will then depend on the preferences of the decision-maker. The reader is referred to \cite{chugh2017survey} for a comprehensive survey on \emph{deterministic} multi-objective methods for engineering problems. In such methods, it is usually assumed that the objective and, if applicable, constraint functions can be observed without noise (e.g., through a deterministic simulator, or noise-free experiments). However, in the adhesive bonding problem discussed here (and in practical settings in general), the objective and constraint observations are \emph{noisy} \citep{Forr09,horn2017first}.

\subsection{Adhesive bonding process: problem setting}
\label{sec:problem}

\begin{figure}[hbt!]
\center
\includegraphics[width=11.5cm]{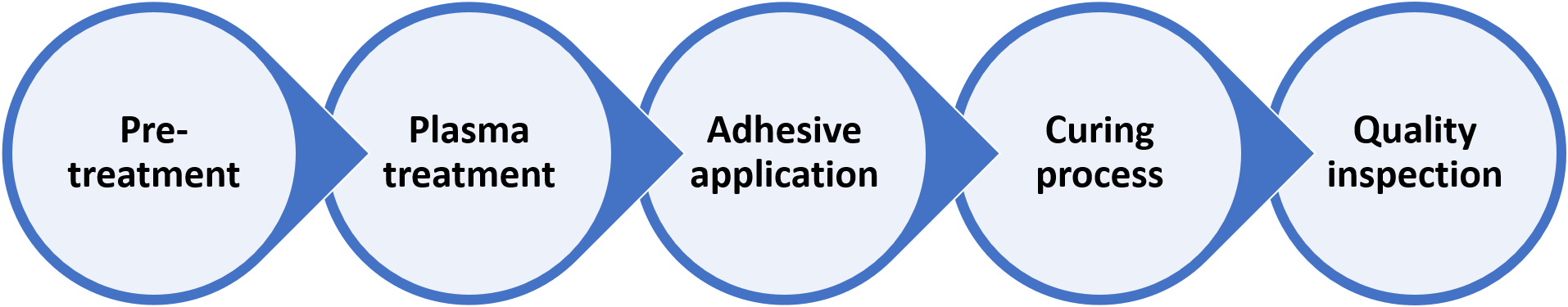} 
\caption{Schematic overview of the adhesive bonding process }
\label{fig:abp}
\end{figure}

\noindent The bonding process we focus on joins two PolyPhenylene Sulfide (PPS) substrates using Araldite 2011 adhesive. Figure \ref{fig:abp} shows the general procedure of the adhesive bonding process. 

\noindent The optimization focuses on the Plasma treatment step. The plasma treatment chemically modifies the top surface layer of the PPS substrate so that the surface energy increases, which impacts the adhesion strength (i.e., the strength of the connection between the adhesive and the substrate). In this process, six parameters play a role: (1) whether the surface is pre-processed or not (cleaning to remove dust and grease, which may prevent a good connection between the adhesive and the substrate), (2) the power setting of the plasma torch, (3) the speed at which the plasma torch moves across the samples, (4) the distance between the plasma torch nozzle and the sample, (5) the number of passes of the plasma torch over the sample, and (6) the time between the plasma treatment and the application of the glue (as the plasma effect reverses over time). The adhesion strength is very sensitive to the configuration of these parameters.  

\begin{figure}[!htbp]
\centering
\begin{subfigure}[b]{0.24\textwidth}
\centering
\includegraphics[width=2.5cm, height=4.5cm]{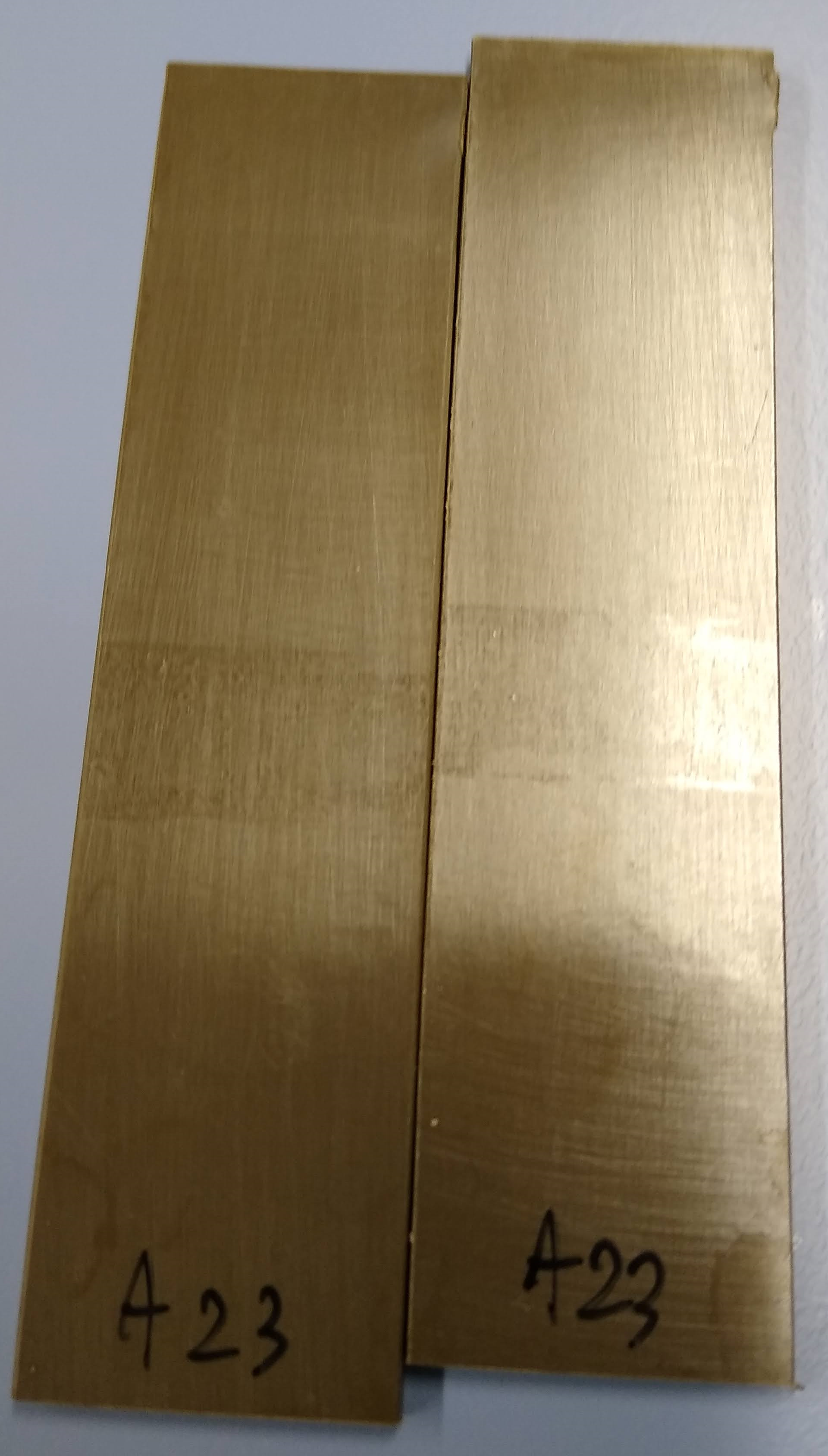} 
\caption{Initial substrate}
\label{fig:ss}
\end{subfigure}
\hfill
\begin{subfigure}[b]{0.24\textwidth}
\centering
\includegraphics[width=2.5cm, height=4.5cm]{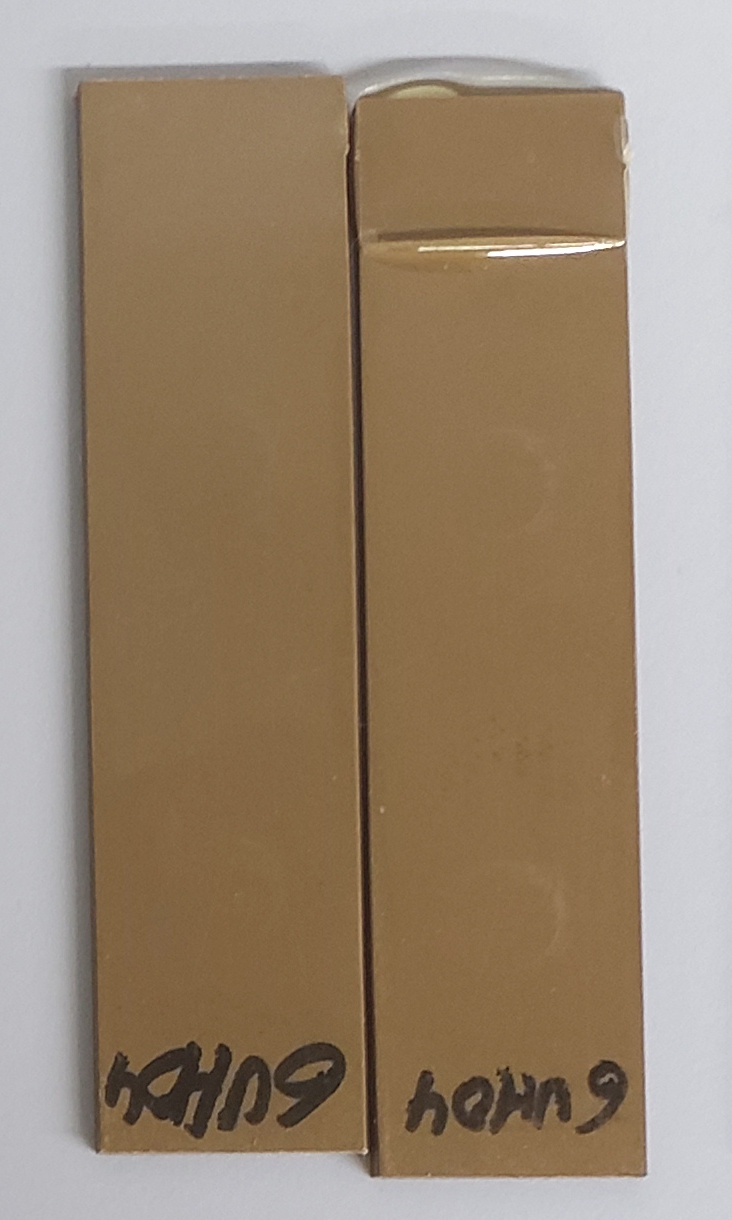} 
\caption{Adhesion failure}
\label{fig:af}
\end{subfigure}
\hfill
\begin{subfigure}[b]{0.24\textwidth}
\centering
\includegraphics[width=2.5cm, height=4.5cm]{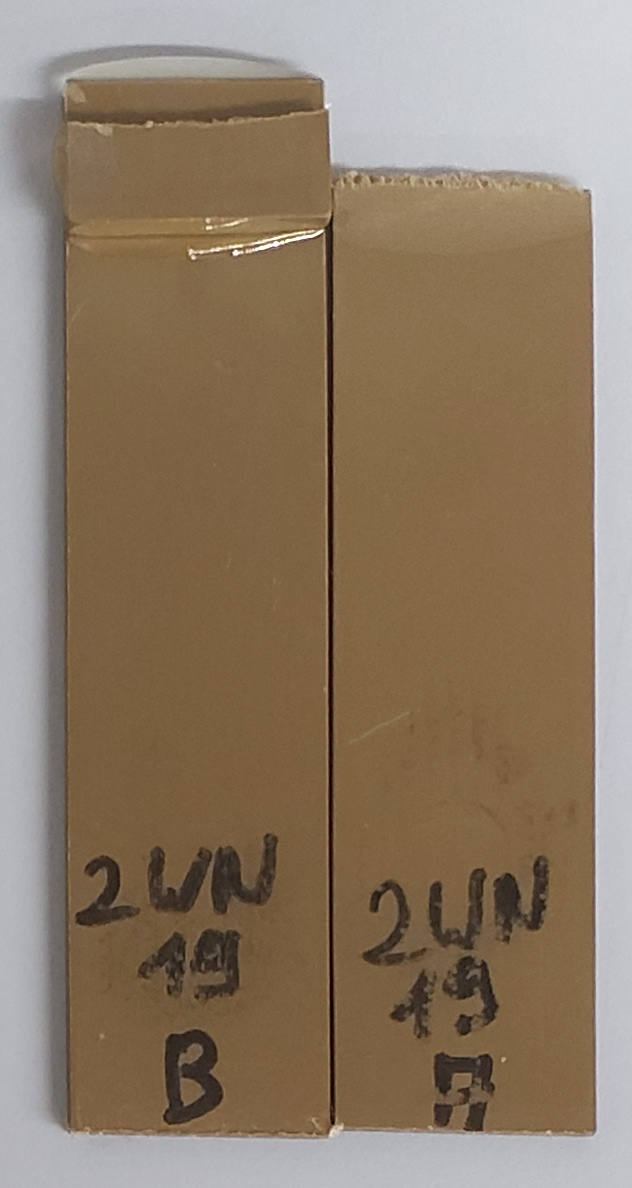} 
\caption{Substrate failure}
\label{fig:sf}
\end{subfigure}
\hfill
\begin{subfigure}[b]{0.24\textwidth}
\centering
\includegraphics[width=2.5cm, height=4.5cm]{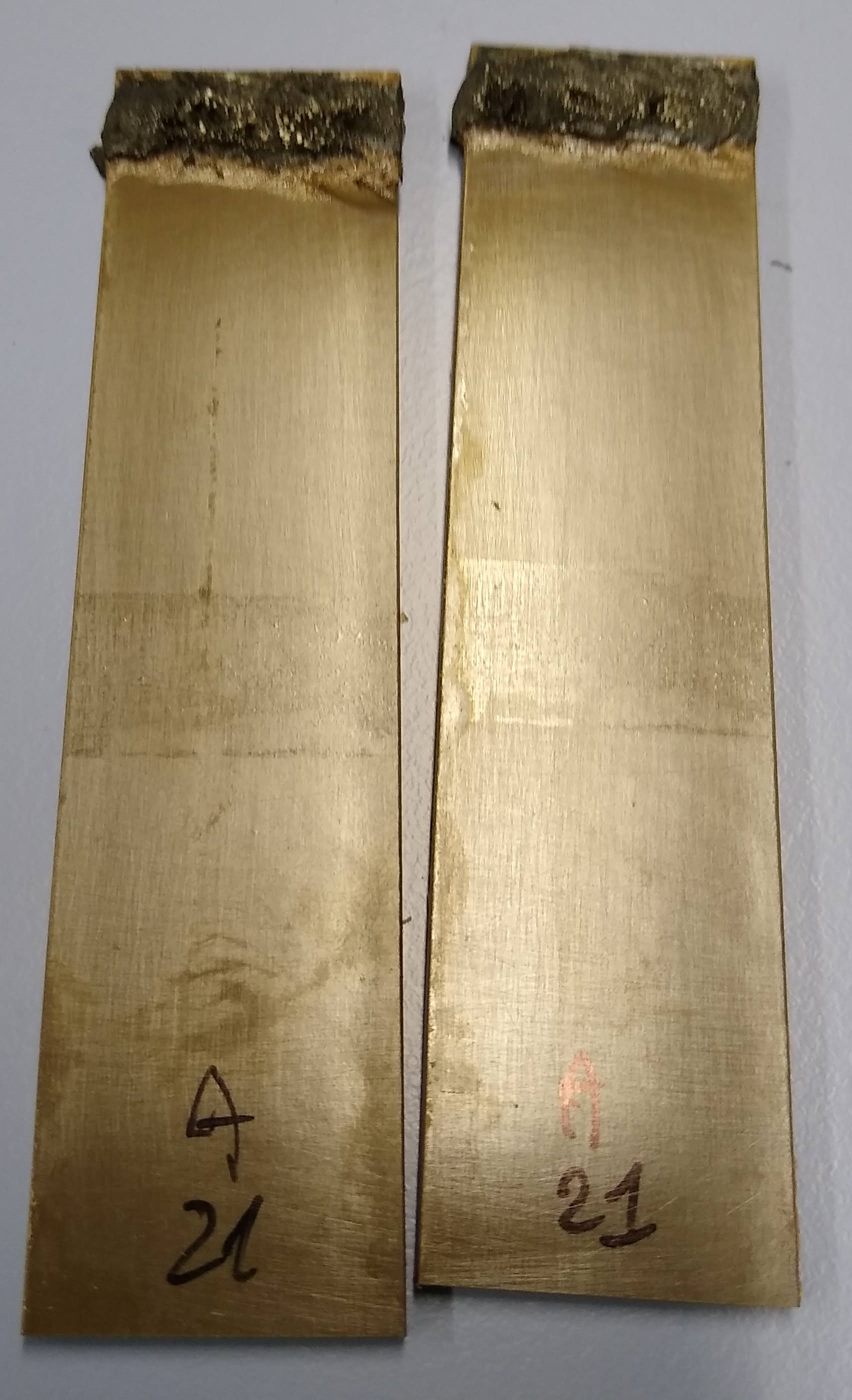}
\caption{Burned sample}
\label{fig:vd}
\end{subfigure}
\caption{Illustration of failure modes that may result from the stress test applied to a sample. The adhesive is applied on the substrates (a) and the failure mode can be either adhesion failure (b), substrate failure (c) or cohesive failure (not shown). In addition, visual damage might be observed after the experiment; e.g., burned sample (d) }
\label{fig:samples}
\end{figure}

Using lab experiments, stress tests can be performed to check the outcomes of samples that have been treated with any particular plasma parameter configuration: the lap shear strength of the sample (MPa), the failure mode (adhesive, cohesive, or substrate failure), the production cost of the sample (in euros), and the potential occurrence of visual damage (the substrates may burn when heated above their maximum allowable temperature during plasma treatment). Such physical experiments are expensive, as they require the whole process in Figure \ref{fig:abp} to be performed, involving a human operator. Figure \ref{fig:samples} shows different failure modes and an example of visual damage.

The goal of the optimization is to set the plasma process parameters in such a way that (1) the tensile strength (TS) is maximized, (2) the production cost (PC) is minimized, and (3) adhesive failures and visual damage are avoided. As objectives (1) and (2) are in conflict, this is a \emph{multi-objective} optimization problem. The goal is to find a set of solutions that reveal the essential trade-offs between these objectives (i.e., those solutions for which no objective can be improved without negatively affecting the performance of any other objective) while meeting the constraints (3). Equation \ref{eq:abp_moo} formally defines this optimization problem as

\begin{mini}|s|
{}{\left [-TS(\mathbf{x}), PC(\mathbf{x}) \right] }{\label{eq:abp_moo}}{}
\addConstraint{0.5 - \text{PoF}(\mathbf{x}) \leq 0}
\end{mini}

\noindent where the notation $\text{PoF}(\mathbf{x})$ refers to the \emph{probability of feasibility} of any given process configuration $\mathbf{x}$ (estimated as the fraction of replications in which the configuration resulted in a feasible outcome). As the performance evaluation is expensive, the optimization algorithm should be able to detect (nearly) Pareto-optimal solutions within a small number of experiments required; collecting large amounts of experimental data is simply financially infeasible. In the following section, we discuss how the use of Bayesian optimization allows us to develop such a data-efficient optimization approach.

\section{Constrained Bayesian multiobjective optimization: proposed algorithms}\label{sec:moo-meth}

BO is a supervised learning technique, that starts with the evaluation of an initial set of input points with good space-filling properties (obtained, e.g., by means of Latin hypercube sampling or quasi-random sequences). The BO literature suggests setting the number of initial design points equal to $k=10d$ (with $d$ the number of dimensions of the input space), but smaller designs have also been advocated \citep{Loeppkyetal2009,zeng2022efficient}. Next, it constructs a \emph{surrogate model} (also referred to as \emph{metamodel}), describing our belief about the response functions under study, based on the samples observed so far. Further observations are then added sequentially using an \emph{acquisition function} that chooses the most promising input point (referred to as \emph{infill point}) as the one to be evaluated next. This acquisition function should balance exploration (in sparsely observed regions of the input space), and exploitation (in regions that are already known to contain good values).  The metamodels are then updated, and the algorithm continues until a stopping criterion is met (e.g., the computational budget is depleted).

\begin{figure}[!htbp]
\centering
\includegraphics[width=\textwidth]{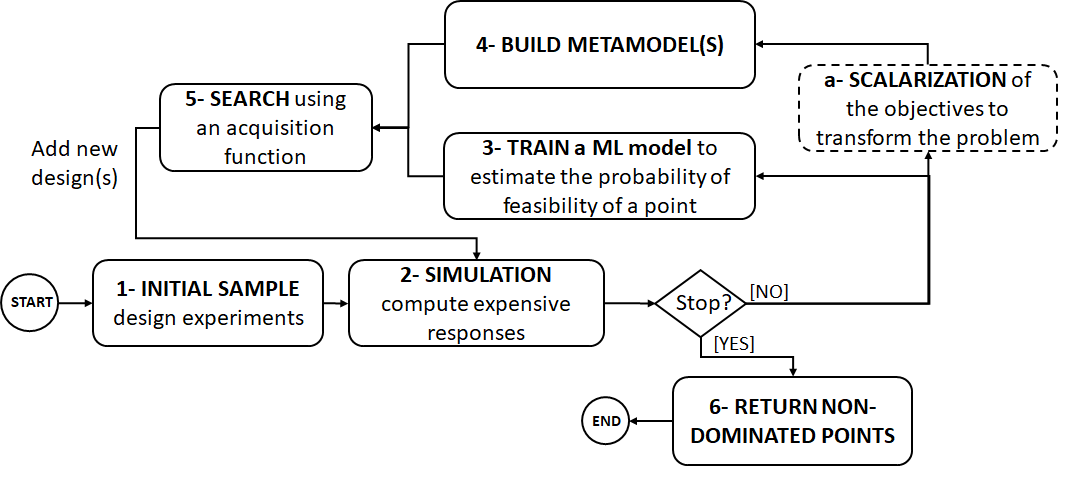} 
\caption{Constrained Bayesian multi-objective optimization algorithms: general steps.}
\label{fig:algorithm}
\end{figure}

Each BO algorithm thus has two key elements: the type of metamodel used, and the type of acquisition function. Several acquisition functions exist (see \cite{frazier2018bayesian} and \cite{Seb19rev} for single and multiobjective surveys respectively); allegedly, the \emph{expected improvement (EI)} remains one of the most commonly used ones in practice \citep{jones1998efficient}, and GP regressors are standard metamodels in the BO literature \citep{williams2006gaussian}. In this work we propose two different algorithms for constrained multi-objective optimization using EI: cMEI-SK, and cEHVI-SK. Both methods follow the same general steps depicted in Figure \ref{fig:algorithm}, but differ in the acquisition function used: while cMEI-SK uses a scalarization approach to transform the problem into multiple single-objective ones (a.o., \cite{knowles2006parego,Seb19}), cEHVI-SK uses the \emph{Expected Hypervolume Improvement (EHVI)} acquisition function \citep{emmerich2011hypervolume,couckuyt2014fast}.

These two acquisition functions are well-known in the BO literature. However, their application has mostly been reported in deterministic unconstrained settings \citep{zhan2020expected}. A few extensions have been proposed in the literature to handle constraints (e.g., \cite{feliot2017bayesian}), and to handle observational noise (e.g., \cite{Seb19,daulton2021parallel}). Given that in our problem setting the feasibility of a process configuration is evaluated with physical tests, where both the objectives and constraints not only are noisy but also of different nature (see Section \ref{sec:problem}), such methods cannot be used out-of-the-box. Further details on the proposed algorithms (and thus our main contributions) are given in sections \ref{subsec: feas}, \ref{subsec: MEI algo} and \ref{subsec: EHVI algo}. Moreover, we explicitly differentiate two types of GP models: \emph{ordinary kriging (OK)} \citep{jones1998efficient,kleijnen2018design} and \emph{stochastic kriging (SK)} \citep{Anke10}. While any GP model can accommodate noisy evaluations \citep{williams2006gaussian}, OK metamodels are limited to homogeneous noise. The seminal work of \cite{Anke10} extended OK metamodels to handle heterogeneous noise (referred to as SK metamodels); as explained further in subsection \ref{subsec: metamodels}, we exploit the information extracted from both types of metamodels. 

\subsection{Ordinary versus stochastic kriging metamodels}
\label{subsec: metamodels}

In ordinary kriging, it is assumed that the observations of the response function under study are deterministic (i.e., they do not exhibit uncertainty). Assume that we observed this function at $n$ input locations $\mathbf{x}^{(i)}, i=\{1, \dots, n\}$ (contained in matrix $\mathbf{X}$), yielding function values $\mathbf{y}^{(i)}, i=\{1, \dots, n\}$ (contained in matrix $\textbf{Y}$). This unknown function $f^{(i)}$ for a point $\mathbf{x}^{(i)}$ is then modeled as

\begin{equation}
\label{eq:regression}
f^{(i)}=m(\mathbf{x}^{(i)}) + M(\mathbf{x}^{(i)})
\end{equation}

In this expression, $m(\mathbf{x})$ represents the mean of the process; it is often supposed to be just a constant ($\beta_0$). The notation $M(\mathbf{x})$ denotes a realization of a Gaussian random field with mean zero, and a covariance structure that exhibits spatial correlation (according to a \emph{covariance function} or \emph{kernel} $k(\cdot, \cdot)$): 

\begin{equation}
    Cov(\mathbf{y}^{(i)}, \mathbf{y}^{(j)})=k(\mathbf{x}^{(i)}, \mathbf{x}^{(j)}),
\end{equation}
Often, this covariance function is assumed to be stationary, meaning that its outcome only depends on the \emph{distance} between the input locations. Multiple kernel functions exist in the literature \citep{williams2006gaussian}; a popular choice is the \emph{squared exponential} or Gaussian kernel:

\begin{equation}
\label{"eq:gaussian_kernel"}
k(\mathbf{x}^{(i)}, \mathbf{x}^{(j)})=\sigma^2 exp\left [ -\sum_{k=1}^d\frac{\left [ x^{(i)}_k-x^{(j)}_k \right ]^2}{2l_k^2} \right ]
\end{equation}

\noindent where $\sigma^2$ is the process variance, and $l_k$ is the length scale of the process along dimension $k$. The parameters $\beta_0$, $\sigma^2$, and $l_k$  are estimated from the already observed data, usually by means of maximum likelihood estimation. 

The ordinary kriging prediction at an arbitrary unobserved location $\mathbf{x}^{(*)}$ is then given by:

\begin{equation}
 \label{eq:ok_p}
    \widehat{f}_{OK}(\mathbf{x}^{(*)}) = \beta_0 + k_*[K_{n}]^{-1}(\textbf{Y}-\mathbf{1}_n\beta_0)
\end{equation}

\noindent where $\mathbf{1}_n$ is a $n \times 1$ vector of ones. The mean squared error on the prediction (MSE, also referred to as \emph{kriging variance}) is given by \citep{jones1998efficient}: 

\begin{equation}
    \widehat{s}_{OK}^2(\mathbf{x}^{(*)}) = k_{**}-k_*K_n^{-1}k_*^{T}
    \label{eq:mse}
\end{equation}

\noindent where

\begin{equation}
K_n = [k(\mathbf{x}^{(i)}, \mathbf{x}^{(j)})], \quad i, j \in \{1, \dots, n\}
\label{eq:Kn}
\end{equation}

\begin{equation}
k_* = [k(\mathbf{x}^{(*)}, \mathbf{x}^{(i)})], \quad i \in \{1, \dots, n\} 
\label{eq:k*}
\end{equation}

\begin{equation}
k_{**} = k(\mathbf{x}^{(*)}, \mathbf{x}^{(*)})
\label{eq:k**}
\end{equation}

Note that the uncertainty involved in the model of Equation \ref{eq:regression} is \emph{imposed} on the problem by assumption, to aid the construction of the predictive model. For this reason, it is also referred to as the \emph{extrinsic uncertainty} \citep{Anke10}. 

In many problem settings, however, the observations of the response function are \emph{not} deterministic: evaluating the same input location $\mathbf{x}^{(i)}$ multiple times (i.e., taking multiple replications) results in a \emph{different} observation $\mathbf{y}^{(r,i)}$ in each replication $r$. The observations thus exhibit \emph{noise}, which stems from the experiments (this is also referred to as \emph{intrinsic} noise). In the literature, this noise is often assumed to be \emph{homogeneous} and modeled using OK metamodels (see e.g., \cite{horn2017first}). In practice, however, the structure and level of the noise often depend on the input variables \citep{kleijnen2018design}. \emph{Stochastic kriging} metamodels \citep{Anke10} explicitly account for this intrinsic (input-dependent) noise. It models the observed response value in the \textit{r}-th replication at design point $\mathbf{x}^{(i)}$ as:

\begin{equation}
   f^{(r,i)}=m(\mathbf{x}^{(i)}) + M(\mathbf{x}^{(i)})+\epsilon_r(\mathbf{x}^{(i)})
\end{equation}

\noindent where $m$ and $M$ are defined as in Equation \ref{eq:regression}, and $\epsilon_r(\mathbf{x}^{(i)})$ is the intrinsic noise observed in replication $r$. The SK prediction at an unobserved location $\mathbf{x}^{(*)}$ is then given by

\begin{equation}
 \label{eq:sk_p}
    \widehat{f}_{SK}(\mathbf{x}^{(*)}) = \beta_0 + k_*[K_{n}+\Sigma_{\epsilon}]^{-1}(\textbf{Y}-\mathbf{1}_n\beta_0),
\end{equation}

and the MSE on this prediction is given by

\begin{equation}
    \widehat{s}_{SK}^2(\mathbf{x}^{(*)}) = k_{**}-k_*\left [K_n^{-1}+\Sigma_{\epsilon} \right ]k_*^{T}.
    \label{eq:sk_mse}
\end{equation}

In these expressions, $K_n$, $k_*$ and $k_{**}$ are as defined in Equations \ref{eq:Kn}, \ref{eq:k*} and \ref{eq:k**} respectively, and 

\begin{equation}
\Sigma_{\epsilon}=diag \left [\frac{Var(\mathbf{y}^{(1)})}{r^{(1)}}, \dots, \frac{Var(\mathbf{y}^{(n)})}{r^{(n)}} \right ],
\end{equation}

\noindent where $Var(\mathbf{y}^{(i)})$ refers to the sample variance of the replication outcomes at location $\mathbf{x}^{(i)}$. The values on the diagonal divide this sample variance by the number of replications ${r^{(i)}}$ taken at each location, and thus reflect the estimated variance of the mean outcome at each observed location. 

\subsection{Probability of feasibility}
\label{subsec: feas}

The \emph{probability of feasibility} (PoF) has been well-studied in the BO literature when the constraints are deterministic \citep{Forr09,couckuyt2014fast,zhan2020expected}. In our problem setting, the constraint feasibility is checked physically (meaning that it will not entail visual damage or adhesive failure), which results in a classification problem with two classes: \texttt{YES} for feasible outcomes and \texttt{NO} for infeasible outcomes. Thus, for each process configuration evaluated, we have a discrete target variable $t \in \{0,1\}$, where 1 denotes \texttt{YES} and 0 denotes \texttt{NO}. We are interested in the probability $p(t=1 \mid \mathbf{X})$; that is, the probability of a process configuration being truly feasible conditioned on the data collected so far. 

To do this, we can train a \emph{classification} model in Step 3 of Algorithm \ref{fig:algorithm}. However, training a binary classifier here is not very helpful, since the \emph{noise} on the constraints will affect the performance for the same input configuration (i.e., replicating the same process configuration may yield different outcomes for constraint violation). For instance, if the same input is replicated 5 times, yielding 2 \texttt{YES} and 3 \texttt{NO}, it is unlikely that training the classifier with e.g., the mode of the outcomes will yield accurate predictions. Therefore, we opt for training a regression model on the proportion of successful outcomes instead: for a given process configuration $\mathbf{x}^{(*)}$, we take the proportion of successful outcomes of the binary target $t$ as

\begin{equation}
    \bar{t}^{(*)} = \frac{\sum_{j=1}^{r^{(*)}} t_j}{r^{(*)}}
    \label{eq:prop}
\end{equation}

\noindent where $r$ is the number of replications. We then approximate the \emph{probability of feasibility} (PoF) of the unobserved locations by fitting an ordinary kriging model (Equation \ref{eq:ok_p}) on the target $\bar{t}$ (denoted $\widehat{f}_c(\mathbf{x})$). That is, given the observed targets $\bar{t} \in \mathbf{T}$ at points $\textbf{X}$, the predicted distribution of an unobserved target $t^{(*)}$ at point $\mathbf{x}^{(*)}$ is given by

\begin{equation}
    \text{PoF}(\mathbf{x}^{(*)}) \simeq \widehat{f}_c(\mathbf{x}^{(*)}) =  p(t^{(*)} = 1 \mid \mathbf{x}^{(*)},\mathbf{X},\mathbf{T}).
    \label{eq:PoF}
\end{equation}

\subsection{cMEI-SK acquisition function}
\label{subsec: MEI algo}

When a scalarization function is used, then only one metamodel is trained on the scalarized objective at each BO iteration. In this work, we use the augmented Tchebycheff scalarization function, which is popular in general multi-objective optimization problems due to the theoretical guarantees it provides \citep{Miet99,knowles2006parego}. The scalarized objective is given by:

\begin{equation}
  Z_{\pmb{\lambda}}(\mathbf{x}) = \max_{j=\{1, \dots, m\}} \lambda_j f_j(\mathbf{x})+\rho \sum_{j=1}^{m} \lambda_j f_j(\mathbf{x})
  \label{eq:scal}
\end{equation}

\noindent where $\pmb{\lambda} = [\lambda_1,...,\lambda_m], \sum_{j=1}^{m} \lambda_j=1, \forall j \in \{1, \dots, m\} $, and $\rho$ is a small positive value (e.g., $\rho=0.05$). We then fit a SK metamodel on $Z_{\pmb{\lambda}}(\mathbf{x})$, explicitly accounting for the intrinsic noise on the scalarized objective (see Equation \ref{eq:sk_p}). The metamodel information is then used in the acquisition function as

\begin{equation}
    \label{eq:ei_gp}
    \begin{aligned}
        \text{MEI-SK}(\mathbf{x}) = &\left [ \widehat{Z}_{SK}(\mathbf{x}_{\min})-\widehat{Z}_{SK}(\mathbf{x}) \right ]\Phi \left [ \frac{\widehat{Z}_{SK}(\mathbf{x}_{\min})-\widehat{Z}_{SK}(\mathbf{x})}{\widehat{s}_{OK}(\mathbf{x})} \right ] \\
        &+\widehat{s}_{OK}(\mathbf{x})\phi \left [ \frac{\widehat{Z}_{SK}(\mathbf{x}_{\min})-\widehat{Z}_{SK}(\mathbf{x})}{\widehat{s}_{OK}(\mathbf{x})} \right ]        
    \end{aligned}
\end{equation}

\noindent where $\widehat{Z}_{SK}(\mathbf{x}_{\min})$ is the SK prediction for the scalarized function ($Z_{\pmb{\lambda}}$) at $\mathbf{x}_{\min}$ (i.e., the point having the lowest sample mean for the scalarized objective among all feasible points already sampled), and $\phi(\cdot)$ and $\Phi(\cdot)$ are the standard normal density and standard normal distribution function, respectively \citep{quan2013simulation,Seb19}. For an arbitrary input location, the corresponding constrained MEI (denoted cMEI-SK) is given by

\begin{equation}
    \text{cMEI-SK}(\mathbf{x}) = \widehat{f}_c(\mathbf{x}) \times \mathrm{MEI}(\mathbf{x})
    \label{eq:cmei}
\end{equation}

\noindent Note that at each BO iteration, a new weight vector $\pmb{\lambda}$ is selected from a set of weights distributed uniformly, allowing the algorithm to sample points across the entire Pareto front \citep{knowles2006parego}.

\subsection{cEHVI-SK acquisition function}
\label{subsec: EHVI algo}

When scalarization is not used, an independent SK metamodel is trained for each of the $m$ objectives. The Expected Hypervolume Improvement (EHVI) is a popular acquisition function in \emph{unconstrained} and \emph{deterministic} settings \citep{loka2022bi,zhan2020expected}. The hypervolume is the size of the space dominated by a Pareto front \textbf{P} given a reference point \citep{zitzler1998multiobjective}. Therefore, the hypervolume improvement of an objective vector $\mathbf{y} \in \mathbb{R}^m$ is defined as the increment of the hypervolume indicator after $\mathbf{y}$ is added to the current approximation of \textbf{P} \citep{emmerich2006single,couckuyt2014fast} 

\begin{equation}
    I(\mathbf{y, \mathbf{P}}) = \mathcal{H}(\mathbf{P}  \cup \{\mathbf{y}\}) - \mathcal{H}(\mathbf{P})
    \label{eq:hi}
\end{equation}

\noindent where $\mathcal{H}$ is the hypervolume calculation function. The EHVI can be defined as the integration of the hypervolume improvement function over the non-dominated area using the metamodel prediction \citep{emmerich2011hypervolume}

\begin{equation}
    \label{eq:ehvi}
        \text{EHVI-SK}(\mathbf{x})= \int_{\mathbf{y}(\mathbf{x}) \in A}I(\mathbf{y}(\mathbf{x}), \mathbf{P})  \prod_{i=1}^{m}\frac{1}{\widehat{s}_{SK_i}(\mathbf{x})}\phi \left (\frac{y_i(\mathbf{x})-\widehat{f}_{SK_i}(\mathbf{x})}{\widehat{s}_{SK_i}(\mathbf{x})} \right) \; dy_i(\mathbf{x})
\end{equation}

\noindent where $A$ stands for the non-dominated area and $\phi(\cdot)$ is the standard normal density distribution function. The terms $\widehat{f}_{SK_i}(\mathbf{x})$ and $\widehat{s}_{SK_i}(\mathbf{x})$ represent the objective and uncertainty estimators of the stochastic GP model respectively. Previous studies have already used EHVI, often assuming noiseless objectives \citep{couckuyt2014fast,daulton2020differentiable,loka2022bi}, or with homogeneous noise at best \citep{daulton2021parallel,koch2015efficient}. Finally, for a novel input configuration $\mathbf{x}^{(*)}$, the corresponding constrained EHVI (denoted cEHVI-SK) is given by

\begin{equation}
    \text{cEHVI-SK}(\mathbf{x}^{(*)}) = \widehat{f}_c(\mathbf{x}^{(*)}) \times \text{EHVI-SK}(\mathbf{x}^{(*)}) 
    \label{eq:cehvi}
\end{equation}

\section{Design of experiments}
\label{sec:experiment_design}

To test the proposed optimization approach, we benchmark its performance against five state-of-the-art constrained EMOAs: C-NSGA-II \citep{brownlee2015constrained}, C-MOEA/D \citep{jain2013evolutionary}, C-TAEA \citep{li2018two}, C-MOPSO \citep{gao2012constrained}, an adaptation of C-NSGA-II \citep{qin2019bayesian} to use the OK metamodel prediction to generate new populations (OK-C-NSGA-II), and a modification to the surrogate-assisted evolutionary algorithm K-RVEA \citep{chugh2016surrogate}, which we refer to as C-K-RVEA. The latter uses GP surrogates to approximate the objective functions, and it was modified to handle infeasible configurations with a penalization factor given by PoF (see Equation \ref{eq:abp_moo}). To evaluate the impact of SK, we also implemented our proposed approaches using OK models; these are denoted with an `OK' in the algorithm name. Furthermore, \cite{zhan2020expected} present a different formulation for constrained EI by using the definition of PoF presented in \cite{sobester2008engineering}. However, we observed in our experiments that the formulation we propose in Equation \ref{eq:cmei} got, on average, superior results compared to the results obtained with the formulation suggested in \cite{zhan2020expected}. Appendix \ref{ap:cei} details these results.

The optimization of the acquisition function in Bayesian optimization tends to be non-trivial, as the function is often non-linear, non-convex, and multimodal \citep{NIPSaf}. Here we use a \emph{Particle Swarm Optimization} (PSO) algorithm to find the infill point that maximizes the proposed acquisition functions (i.e., the fitness function of this inner optimization). Our choice is motivated by the good performance and low computational time observed in other studies with high-dimensional search space \citep{Yarat2021}. With PSO, the position of the particle represents the values of each variable to optimize.

Since the physical experiments are very expensive, a Matlab process simulator was provided by the Joining \& Materials Lab\footnote{\url{https://www.flandersmake.be}}. This simulator predicts the lap shear strength of the sample (MPa), failure mode (adhesive, substrate, or cohesive failure), sample production cost (in euros), and visual quality outcome (\texttt{YES} in a matter of seconds. It is critical to note that this simulator is \emph{not} meant to be a digital twin of the true process, but rather a tool for the \emph{relative} comparison of the performance of the algorithms under different conditions, at almost zero cost. Table \ref{tab:process_variables} shows the range of each process parameter considered in the optimization problem, and Table \ref{tab:experiments} summarizes the parameters of the optimization algorithms.

\begin{table}[!hbtp]
\centering
\caption{Range of the process settings (input variables) considered in the optimization. }
\label{tab:process_variables}
\begin{tabular}{lp{6cm}cc}
\hline
\multicolumn{1}{c}{\textbf{ID}} & \multicolumn{1}{c}{\textbf{Variable}} & \multicolumn{1}{c}{\textbf{Min}} & \multicolumn{1}{c}{\textbf{Max}} \\ \hline
v1 & Pre-processing & \multicolumn{2}{c}{Yes or No} \\
v2 & Power setting (W) & 300 & 500 \\
v3 & Torch speed (mm/s) & 5 & 250 \\
v4 & Distance between the torch and the sample (cm) & 0.2 & 2 \\
v5 & Number of passes & 1 & 50 \\
v6 & Time between plasma treatment and glue application (min) & 1 & 120 \\ \hline
\end{tabular}
\end{table}

The simulator allows the analyst to experiment with different levels of noise. In real life, multiple factors cause noise to occur. One of these is the so-called \emph{contact angle}\footnote{Other noise factors could not be controlled in the simulator, so they are not further discussed.}: this typical measure reflects the extent to which the adhesive can maintain good contact with the material. This is important to achieve a strong adhesive bond. The contact angle depends on the type of material but also on impurities or contaminants such as wax, oil, plasticizers, etc. present on the material surface. Even though all samples in our setting are made of the same material, variations in the degree of these contaminants occur across samples, implying variations in contact angle. These result in noisy measurements of the final break strength of the bonded joints. A realistic value for the \emph{standard deviation of the contact angle} is $\gamma=30\%$ of the mean, which is what we use in our experiments.

\begin{table}[!hbtp]
\centering
\caption{Summary of the parameters of the optimization algorithms}
\label{tab:experiments}
\begin{tabular}{p{4cm}p{2cm}p{4cm}}
\hline
\multicolumn{1}{c}{\textbf{Setting}} & \multicolumn{1}{l}{\thead{\textbf{BO} \\ \textbf{algorithms}}} & \multicolumn{1}{c}{\thead{\textbf{Evolutionary} \\\textbf{algorithms}}} \\ \hline

Size of initial design/population &  \multicolumn{2}{c}{LHS: $N=20$} \\ \hline 
Crossover probability &   \multicolumn{1}{c}{\textbf{-}} & 0.5 (C-MOEA/D); 0.9 o.w  \\ \hline 
Mutation probability &   \multicolumn{1}{c}{\textbf{-}} & \multicolumn{1}{c}{$0.1$} \\ \hline 
Reference directions & \multicolumn{1}{c}{\textbf{-}} & 19 (C-TAEA); 3 (C-MOEA/D) \\ \hline
Inertia weight & \multicolumn{1}{c}{\textbf{-}} & \multicolumn{1}{c}{0.4 (C-MOPSO)} \\ \hline
C1 factor & \multicolumn{1}{c}{\textbf{-}} & \multicolumn{1}{c}{2 (C-MOPSO)} \\ \hline
C2 factor & \multicolumn{1}{c}{\textbf{-}} & \multicolumn{1}{c}{2 (C-MOPSO)} \\ \hline
Max velocity (\%) & \multicolumn{1}{c}{\textbf{-}} & \multicolumn{1}{c}{5 (C-MOPSO)} \\ \hline
Reference vectors & \multicolumn{1}{c}{\textbf{-}} & \multicolumn{1}{c}{151 (C-K-RVEA)} \\ \hline
Replications &  \multicolumn{2}{c}{$r=5$} \\ \hline

Iterations & \multicolumn{1}{c}{40} & \multicolumn{1}{c}{2} \\ \hline

Acquisition function & \multicolumn{1}{c}{$\text{cMEI-SK} / \text{cEHVI-SK}$}  & \multicolumn{1}{c}{\textbf{-}} \\ \hline

Acquisition function optimization & \multicolumn{1}{c}{PSO*} & \multicolumn{1}{c}{\textbf{-}} \\ \hline

Kernel (for all GPR models) & \multicolumn{1}{c}{Gaussian}  & \multicolumn{1}{c}{\textbf{-}} \\ \hline

\multicolumn{3}{l}{\makecell[tl]{ * PSO configuration: swarm size = 50, max iterations=1800, max stall \\iterations = 10, tolerance = $1e^{-6}$ }} \\
\end{tabular}
\end{table}

Given the experimental setting in Table \ref{tab:experiments}, each algorithm evaluates exactly 60 process configurations in an expensive way, with 5 replications per configuration (i.e., 300 expensive evaluations in total). The BO algorithms start with an initial design of 20 process configurations (note that this is smaller than the usual choice of $k=11d-1$); exactly one infill point is then added in each of the following 40 iterations. The EMOAs, by contrast, use an initial population of 20 process configurations (the same initial set used by BO algorithms) and in each of the $2$ successive generations, a novel population is generated. As common in the literature, the fitness of the configuration outcomes is evaluated based on their sample means over the 5 replications (note that, by doing so, these algorithms implicitly ignore the fact that this sample mean is in itself uncertain). Our approach takes into account both the sample mean and the sample variance though, as explained in Section \ref{sec:moo-meth}. While a total budget of 300 evaluations may seem high, it allows us to also study the progress the algorithms would have obtained at lower budgets, as illustrated below.

We evaluate the quality of the resulting fronts using the well-known hypervolume (HV) and IGD+ indicators \citep{LiYao2019acmcs}. The hypervolume is the volume of the objective space that is dominated by the front obtained, w.r.t. a reference point, and the IGD+ is the average smallest inverted generational distance to the closest member of the true Pareto front. The larger the hypervolume and the lower the IGD+, the better the quality of the front obtained. We also include performance analysis of the proposed methods using the \emph{empirical attainment function} \citep{lopez2010exploratory}. As the front obtained by the algorithms may depend on the initial design, we performed 50 macro-replications, each one starting with a different initial design. 

\section{Results}
\label{sec:results}

\begin{figure}[!htbp]
\centering
\includegraphics[width=8cm]{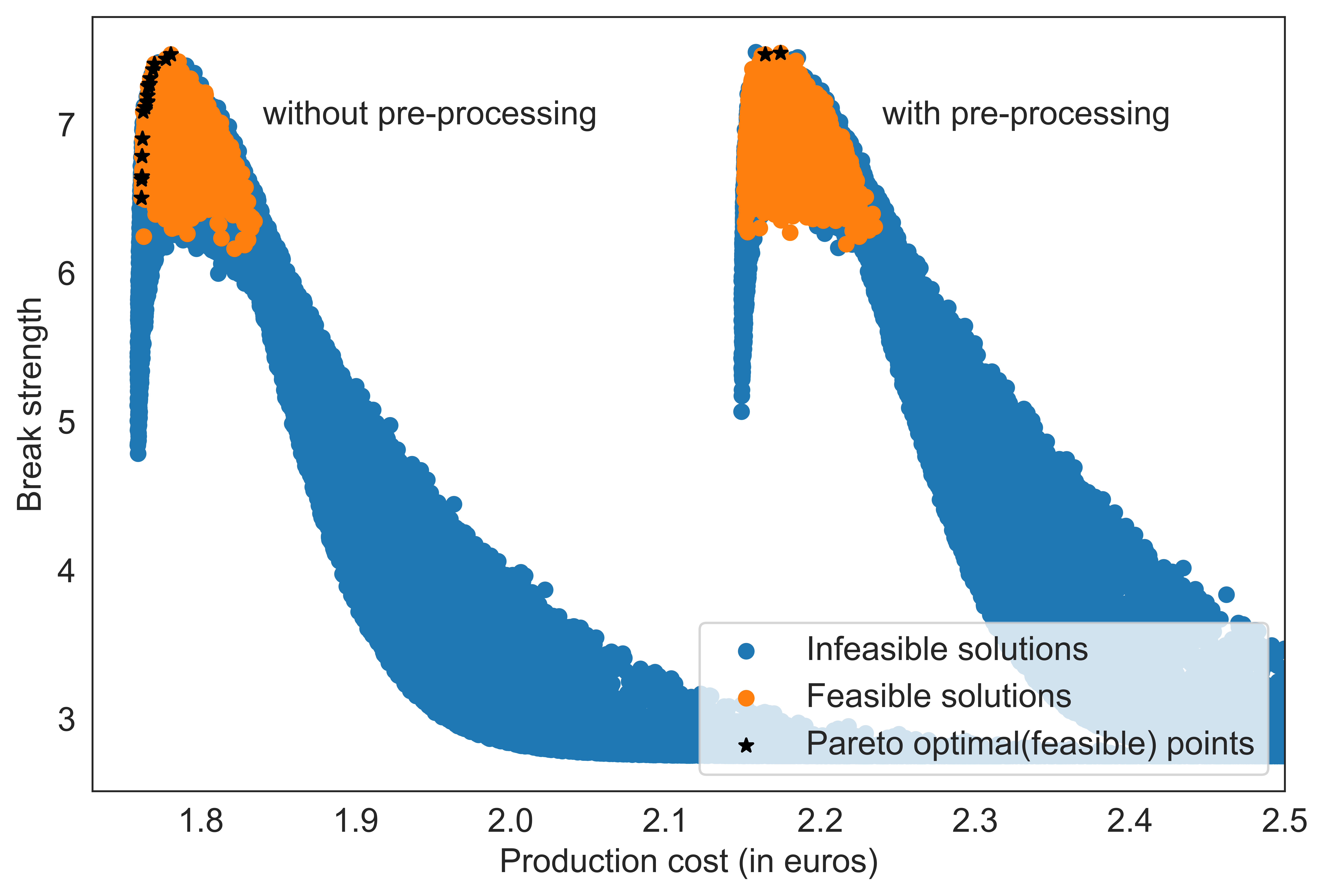} 
\caption{Sample mean of break strength versus production cost, estimated by the simulator for 60 000 random process configurations ($\gamma = 0\%$)}
\label{fig:halton_set}
\end{figure}

To gain some insight into the objective space to be explored by the algorithms under idealized conditions (i.e., if the contact angle could be perfectly controlled), Figure \ref{fig:halton_set} shows the mean responses of the simulator on a set of 60 000 process configurations (with $\gamma = 0\%$). These configurations were determined through Halton sampling, and each configuration was replicated five times. Interestingly, the feasible solutions seem to be clustered in areas with high break strength. Moreover, the use of pre-processing seems to merely lead to a cost increase, while the resulting gains in break strength are very scarce and only minor. The Pareto optimal (feasible) points estimated by means of these Halton results, under these idealized conditions ($\gamma = 0\%$), serve as the benchmark Pareto front to judge the quality of the competing optimization algorithms.

\begin{figure}[!htbp]
\centering
\includegraphics[width=11cm]{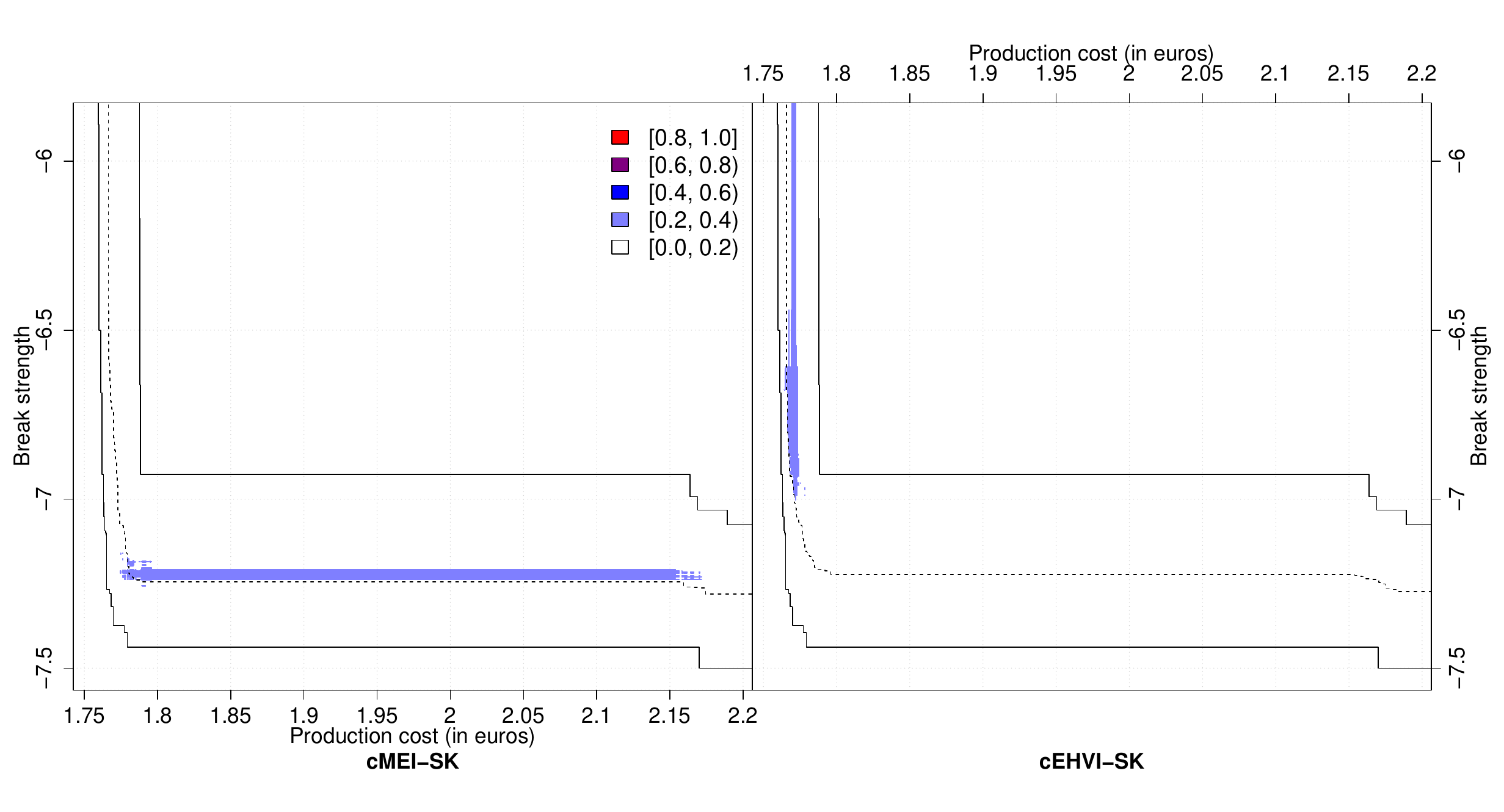}  
\caption{Visualization of the differences between the EAFs using different \textit{Improvement} functions in the BO methods. The Y-axis is inverted to allow for the minimization of both objectives. Each plot shows the median surface (dashed line), along with the best and worst surfaces (full lines). The areas where a better attainment surface is obtained are indicated by a shaded area (the colors indicate the improvement in the probabilities).}
\label{fig:eas}
\end{figure}

Figure \ref{fig:eas} shows an exploratory analysis of the Pareto fronts obtained by each BO algorithm in the different macro-replications. The analysis is performed using the concept of the empirical attainment function (EAF, \citealt{lopez2010exploratory}). For each point, the EAF gives an estimate of the probability that this point is dominated (attained) by the Pareto front put forward by the given algorithm. Connecting points with the same given EAF value yields an \emph{attainment surface} that separates the objective space into two regions: those objective vectors that are attained by the resulting Pareto fronts with (at least) that probability, and those that are not. The attainment surfaces allow us to summarize the location of the objective vectors obtained by a stochastic algorithm. The median attainment surface, for instance, consists of the objective vectors that are attained by half of the runs (representing a probability of 50\%). Similarly, the worst-case results of an algorithm are reflected in the worst attainment surface, whereas the best results are given by the best attainment surface. The shaded areas in the right-hand side of Figure \ref{fig:eas} show the objective areas where the proposed cMEI-SK approach reaches better attainment surfaces minimizing the objective \textbf{cost}. On the other hand, using cEHVI-SK means a better attainment surface minimizing the objective \textbf{break strength}. 

\vspace{-6mm}

\begin{figure}
\centering
\begin{subfigure}[b]{0.45\textwidth}
\centering
\includegraphics[width=4.5cm]{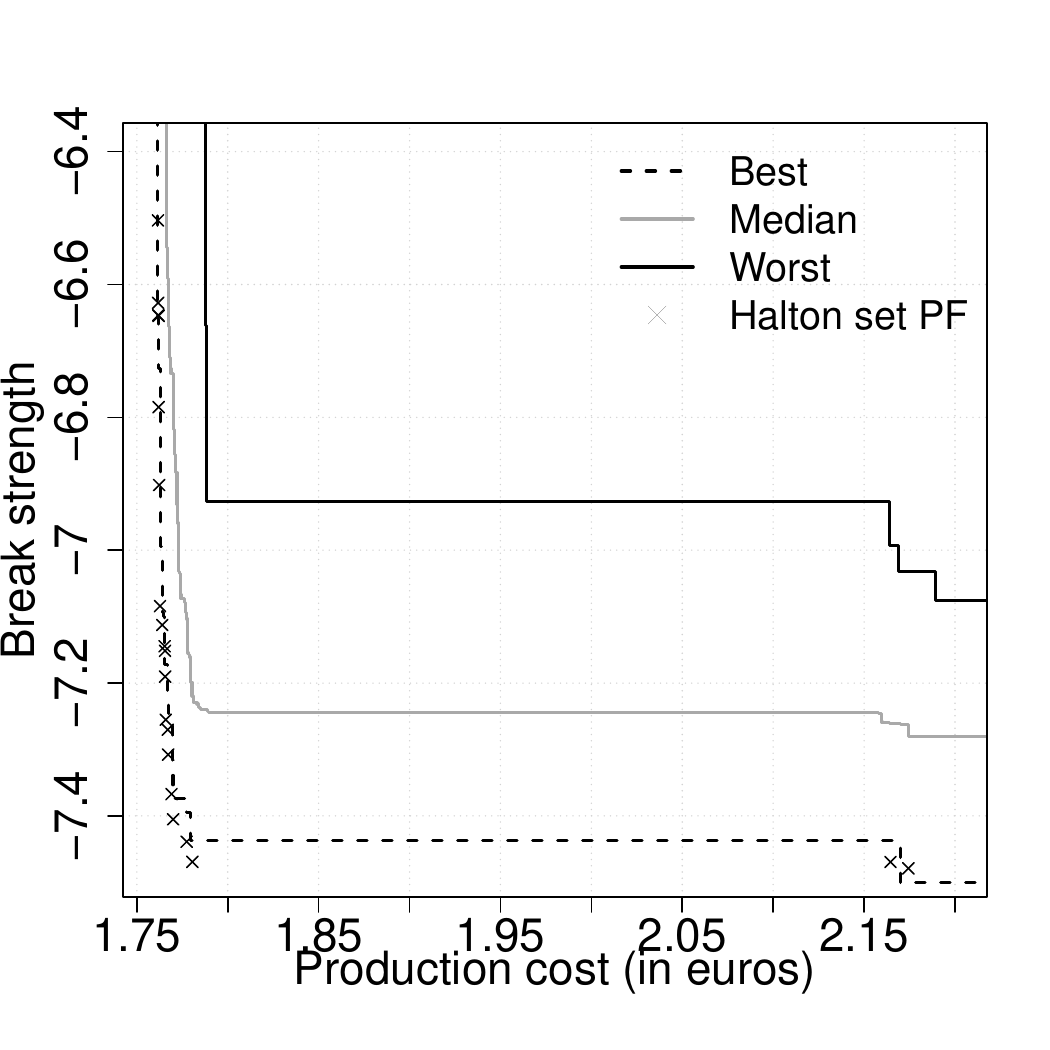}  
\caption{cEHVI-SK}
\label{fig:ehi_H0}
\end{subfigure}
\begin{subfigure}[b]{0.45\textwidth}
\centering
\includegraphics[width=4.5cm]{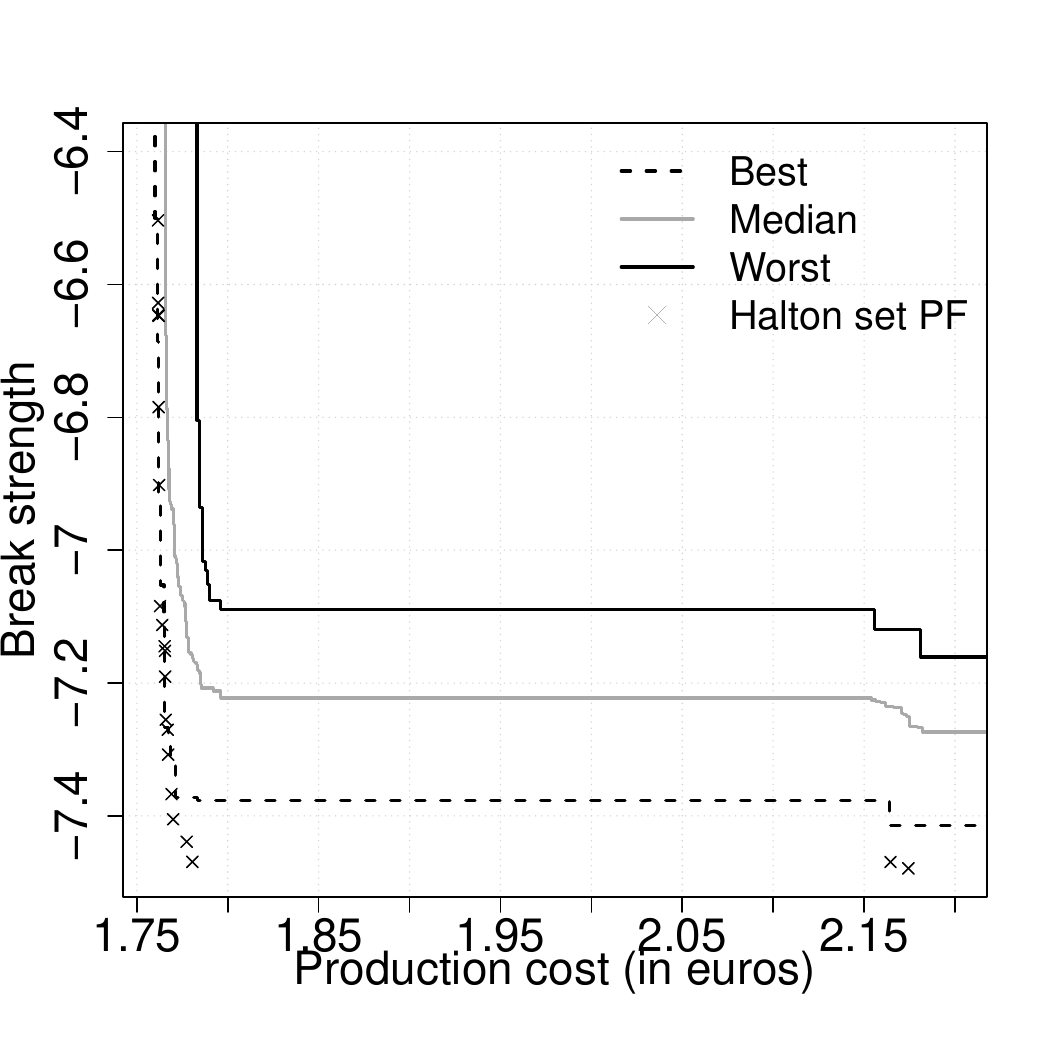}  
\caption{cMEI-SK}
\label{fig:mei_H0}
\end{subfigure}
\caption{Best, median, and worst Pareto front obtained for the BO methods. The Y-axis is inverted to allow minimization of both objectives}
\label{fig:eaf_bo_H0}
\end{figure}

\begin{figure}[!htbp]
\centering
\includegraphics[width=11cm]{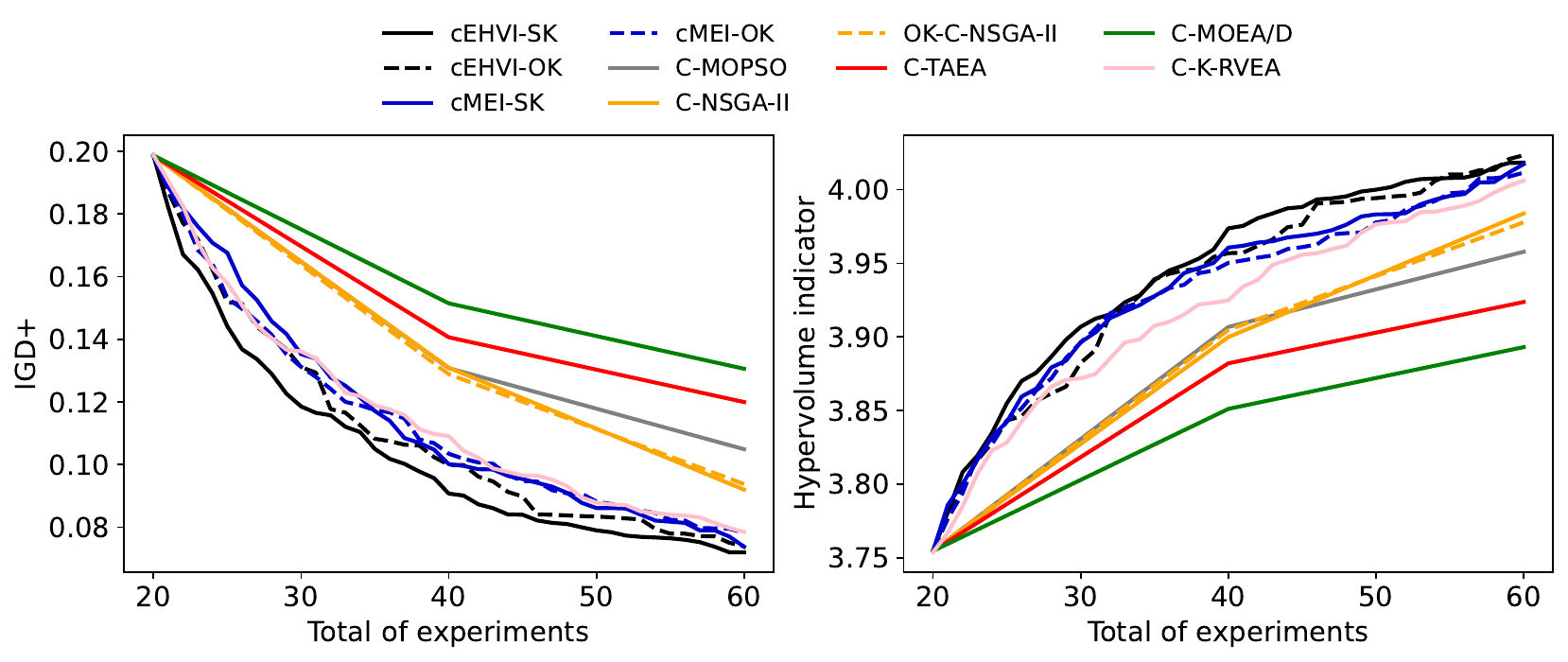}  
\caption{Evolution of the IGD+ metric (Left) and hypervolume indicator (Right) of the Pareto-optimal solutions throughout the search, for $\gamma = 30\%$ (average of 50 macro-replications). The Pareto front obtained from the Halton set is considered to compute the IGD+ metric and the reference point with production cost=3, break strength=4 is used to compute the hypervolume indicator} 
\label{fig:evolution}
\end{figure}

Figure \ref{fig:eaf_bo_H0} shows the final \emph{best} Pareto front obtained by the proposed approaces over 50 macro-replications (along with the \emph{median} and \emph{worst} fronts), for $\gamma = 30\%$. The best Pareto front obtained by cEHVI-SK is very close to the ideal Pareto front estimated by means of the Halton set exploration. It also leads to a \emph{faster} increase in HV, in terms of the number of expensive evaluations performed, than all other algorithms. This is evident from Figure \ref{fig:evolution}, which shows the evolution of the \emph{average} hypervolume and IGD+ obtained (across macro-replications) during the optimization process (again, for $\gamma = 30\%$). This superior performance is also evident from Table \ref{tab:igd}, which shows the results for the average HV, along with those of the average IGD+ (the latter uses the ideal front obtained in the Halton experiment of Figure \ref{fig:halton_set} as the true front). These results highlight that BO methods are able to obtain better quality results for the Pareto front than evolutionary algorithms (including surrogate-assisted ones), particularly at very limited budgets. Appendix \ref{ap:wilcoxon} shows that statistical differences (Wilcoxon test, $\alpha=5\%$), both for HV and IGD+, were mainly focused amongst C-MOPSO, C-MOEA/D, and C-TAEA; and metamodel-based optimization algorithms (including C-K-RVEA).

Overall, ignoring the input-dependent noise associated with the objective \textit{break strength} (thus using the mean value of the replications) negatively influences the performance of optimization algorithms. As shown in Figure \ref{fig:evolution}, the curves of the \emph{deterministic} BO methods (i.e., the proposed algorithms using OK metamodels) are inferior to the one when input-dependency is taken into account. Yet, they remain superior to the evolutionary algorithms, which reinforces the advantages of the exploration/exploitation performed by the optimization of the acquisition function and thus obtaining the best possible trade-off between expected performance and model error. Additionally, we note that surrogate-assisted evolutionary algorithms (such as C-K-RVEA) can indeed benefit from the approximation of the objectives and obtain better configurations than standard EMOAs. However, this performance remains inferior to BO approaches when data-efficiency is needed on top of effective black-box optimization. 

\begin{figure}[!htbp]
\centering
\includegraphics[width=12cm]{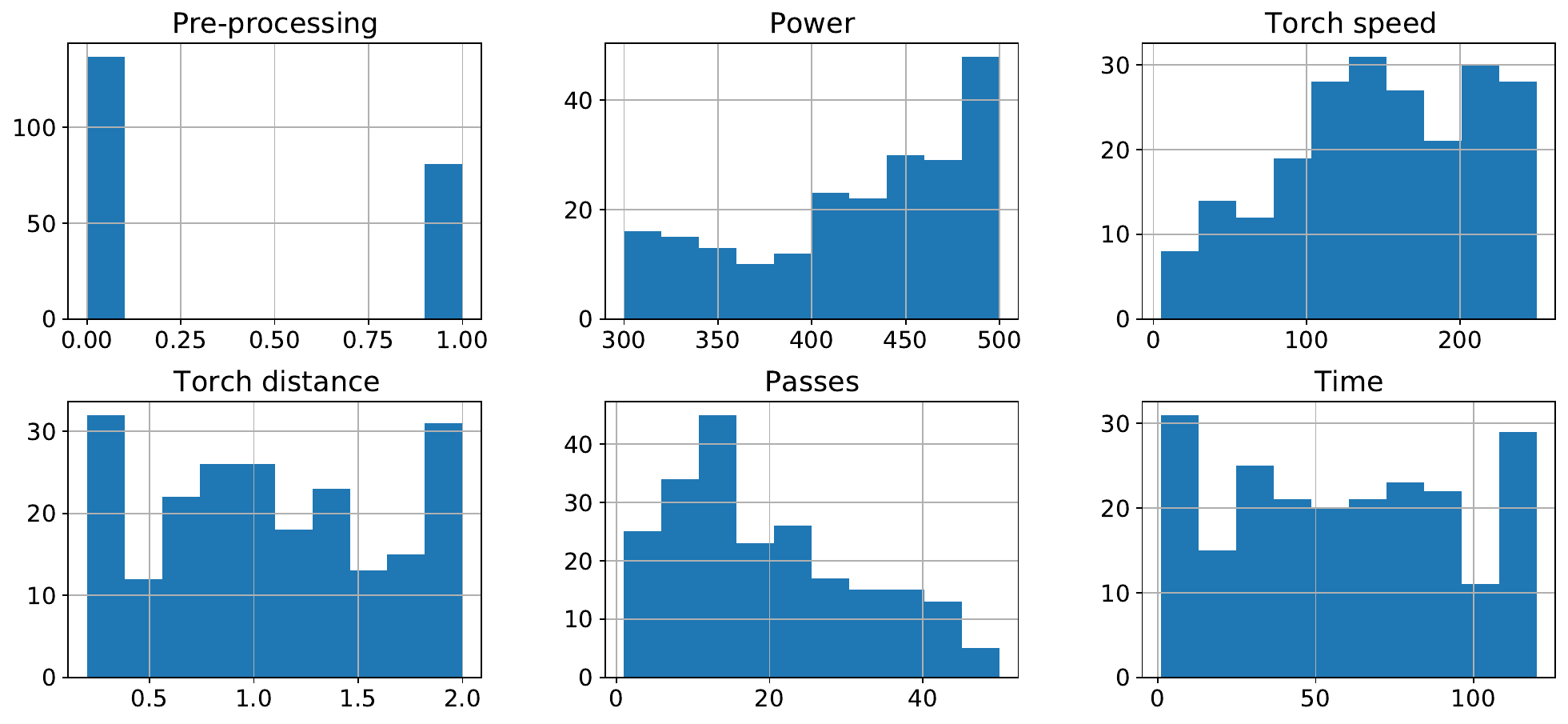}  
\caption{Distribution of the Pareto-optimal input values obtained by cEHVI-SK, across 50 macro-replications.}
\label{fig:dec_space_statistics1}
\end{figure}

Finally, by looking closer at the solutions (in input space) suggested by cEHVI-SK (the one with the highest HV value), we found that $\pm 63\%$ of the solutions skip pre-processing, meaning that the production costs are reduced. Thus in Figure \ref{fig:dec_space_statistics1} we show the distribution of the remaining input variables. Overall, the algorithm suggests that the power should be between 480 W and 500 W, the speed should move at a speed between 127.5 mm/s and 152 mm/s, the distance between the torch and the sample should be in the range of 0.2cm and 0.38 cm, between 11 and 16 passes should be performed, and a time difference between 1 and 13 minutes should be considered before the glue application. 

\begin{table}[!hbtp]
\centering
\caption{Average IGD+ and HV of the fronts obtained over 50 macro-replications, for $\gamma=30\%$}
\label{tab:igd}
\begin{tabular}{lcc}
\hline
& \textbf{IGD+} & \textbf{HV}\\ 
\hline
\textbf{cEHVI-SK} & 0.0719 & 4.0184 \\
\textbf{cEHVI-OK} & 0.0739 & 4.0235 \\
\textbf{cMEI-SK} & 0.0737 & 4.0174 \\
\textbf{cMEI-OK} & 0.0782 & 4.0116 \\
\textbf{C-K-RVEA} & 0.0785 & 4.0062 \\
\textbf{C-NSGA-II} & 0.092 & 3.9839 \\
\textbf{OK-C-NSGA-II} & 0.0937 & 3.979 \\
\textbf{C-MOPSO} & 0.1049 & 3.9579 \\
\textbf{C-TAEA} & 0.1199 & 3.9237 \\
\textbf{C-MOEA/D} & 0.1306 & 3.893 \\ \hline
\end{tabular}
\end{table}

\section{Conclusions}
\label{sec:conclusions}

This paper presented two constrained Bayesian optimization algorithms to solve a bi-objective problem related to the adhesive bonding process of materials (maximizing break strength while minimizing production costs). As the real experiments are carried out physically in a lab are costly, the budget for evaluations is very limited. The proposed Bayesian approach is shown to clearly outperform state-of-the-art evolutionary algorithms, which are commonly used in engineering design when solving general multi-objective, constrained problems. The difference lies in the way the experimental design is guided throughout the search: the Bayesian approach selects infill points based on an (explainable) acquisition function, which is related to the expected merit of the new infill point for optimization. The BO model ensures that the search focuses on infill points that have a high probability of being feasible. Moreover, the GP model used to approximate the objective(s) accounts for the output (heterogenous) noise, whereas the evolutionary algorithms rely simply on the (uncertain) sample means as performance approximations. Moreover, the search in EMOAs (as it is generally with metaheuristic approaches) is guided by hard-to-tune evolutionary operators. The success of evolutionary processes is largely dependent on the availability of a sufficient experimentation budget, which is not always the case in practice.

Our research highlights the superiority of BO methods, particularly using an EHVI-based infill criterion, over traditional EMOAs. We are convinced that the use of Bayesian approaches holds great promise in solving noisy and expensive engineering problems, in terms of both search efficiency (i.e., finding solutions within a limited budget) and search effectiveness (i.e., yielding high-quality solutions). Future research will focus on the further development of an interactive software tool (a first release has been provided and is being tested), allowing lab experts to validate the results generated by the algorithm in a real-life test environment, and on the inclusion of a third objective (minimization of the \textit{debonding} break strength.

\section*{Acknowledgments}
 This work was supported by the Flanders Artificial Intelligence Research Program (FLAIR), the Research Foundation Flanders (FWO Grant 1216021N), and Flanders Make vzw.

\begin{appendices}

\section{Constrained Expected Improvement (CEI)}\label{secA1}
\label{ap:cei}

The Constrained Expected Improvement (CEI,  \citealt{zhan2020expected}) uses one OK metamodel to approximate the expensive objective and one metamodel to approximate each expensive constraint independently. Then, for a constrained optimization problem with $c$ constraints

\begin{mini}|s|
{}{ y(\mathbf{x}), \mathbf{x} \in \mathbb{R} }{\label{eq:moo_extra}}{}
\addConstraint{g_i(\mathbf{x}) \leq 0, \, i=1, 2, \dots, c}
\end{mini}

\noindent the objective value of point $\mathbf{x}$ can be treated as a Gaussian random variable $\mathcal{N}\left ( \widehat{y}(\mathbf{x}), \widehat{s}\mathbf(x) \right )$ and the $i$-th constraint value of $\mathbf{x}$ can be also treated as a Gaussian random variable $\mathcal{N}\left ( \widehat{g}_i(\mathbf{x}), \widehat{e}_i\mathbf(x) \right ), \; i=1, 2, \dots, c$. For this, $\widehat{y}$ and $\widehat{s}$ are the GP prediction and standard error of the objective function respectively, and $\widehat{g}_i$ and $\widehat{e}_i$ are the GP prediction and standard error of the $i$-th constraint function respectively.

Then, from Equation \ref{eq:abp_moo} we can transform the constraint and derive the Probability of Feasibility as:

\begin{equation}
\label{eq:constraint1}
     g(\mathbf{x})=0.5-Pf(\mathbf{x}) \leq 0 = Pf(\mathbf{x}) \geq 0.5 = \frac{Pf(\mathbf{x}) - \widehat{g}(\mathbf{x})}{\widehat{e}(\mathbf{x})} \geq \frac{0.5 - \widehat{g}(\mathbf{x})}{\widehat{e}(\mathbf{x})}
\end{equation}

\begin{equation}
\begin{aligned}
     PoF(\mathbf{x}) = {} & Prob\left ( \frac{Pf(\mathbf{x}) - \widehat{g}(\mathbf{x})}{\widehat{e}(\mathbf{x})} \geq \frac{0.5 - \widehat{g}(\mathbf{x})}{\widehat{e}(\mathbf{x})} \right ) \\
     &= 1-Prob\left ( \frac{Pf(\mathbf{x}) - \widehat{g}(\mathbf{x})}{\widehat{e}(\mathbf{x})} \leq \frac{0.5 - \widehat{g}(\mathbf{x})}{\widehat{e}(\mathbf{x})} \right ) \\
     &= 1-\Phi \left ( \frac{0.5 - \widehat{g}(\mathbf{x})}{\widehat{e}(\mathbf{x})} \right )
\end{aligned}
\end{equation}

In case the GP standing for the objective and constraint function are mutually independent, the CEI can be obtained by combining the EI (to be consistent with our work we used MEI instead and the predictors defined in Equation \ref{eq:sk_p} and Equation \ref{eq:sk_mse}) and PoF as

\begin{multline}
    \text{CMEI-SK}(\mathbf{x}) = \text{MEI}(\mathbf{x}) \times \text{PoF}(\mathbf{x}) \\
    =\left[\left(\widehat{f}_{SK}(\mathbf{x}_{min})-\widehat{f}_{SK}(\mathbf{x})\right)\Phi\left(\mathcal{D} \right)+ \widehat{s}_{OK}(\mathbf{x})\phi\left(\mathcal{D} \right) \right] \\ 
    \times \left[1-\Phi\left(\frac{0.5 - \widehat{g}(\mathbf{x})}{\widehat{e}(\mathbf{x})}\right)\right]
    \label{eq:CEI}
\end{multline}

\noindent and 

\begin{equation} 
\mathcal{D} = \frac{\widehat{f}_{SK}(\mathbf{x}_{min})-\widehat{f}_{SK}(\mathbf{x})}{\widehat{s}_{OK}(\mathbf{x})}
\end{equation}

\begin{figure}[!hbt]
\center
\includegraphics[width=8cm]{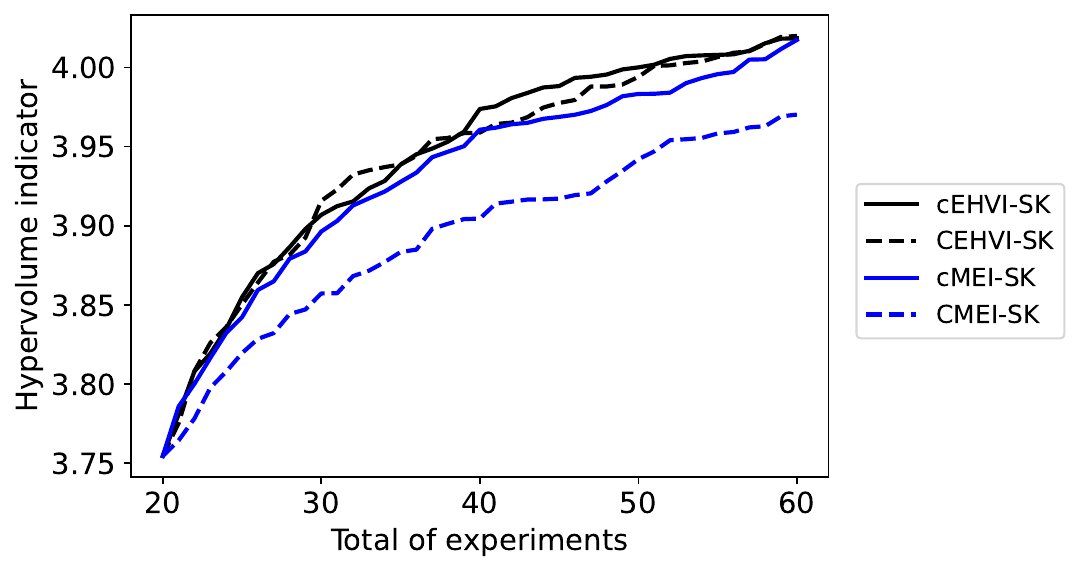} 
\caption{Evolution of the mean hypervolume throughout the optimization (of 50 macro-replications). The reference point [production cost=3, break strength=4] is used to compute this metric}
\label{fig:af_comparison}
\end{figure}

\noindent Note that this equation is different from our proposed Equation \ref{eq:cmei} in the derivation of the PoF. Figure \ref{fig:af_comparison} shows that our proposed cMEI-SK got on average better Pareto fronts (higher hypervolume values) than that of CEI. This suggests that considering the uncertainty predicted by the GP may lead the optimization to points that do not cause an increase in the hypervolume. This was more evident when the improvement was measured with MEI (and fitting only one GP to the scalarized objectives). 

\section{Wilcoxon test results}\label{secA2}
\label{ap:wilcoxon}

\begin{figure}[!h]
\centering
\begin{subfigure}[b]{0.45\textwidth}
\centering
\includegraphics[width=5.5cm]{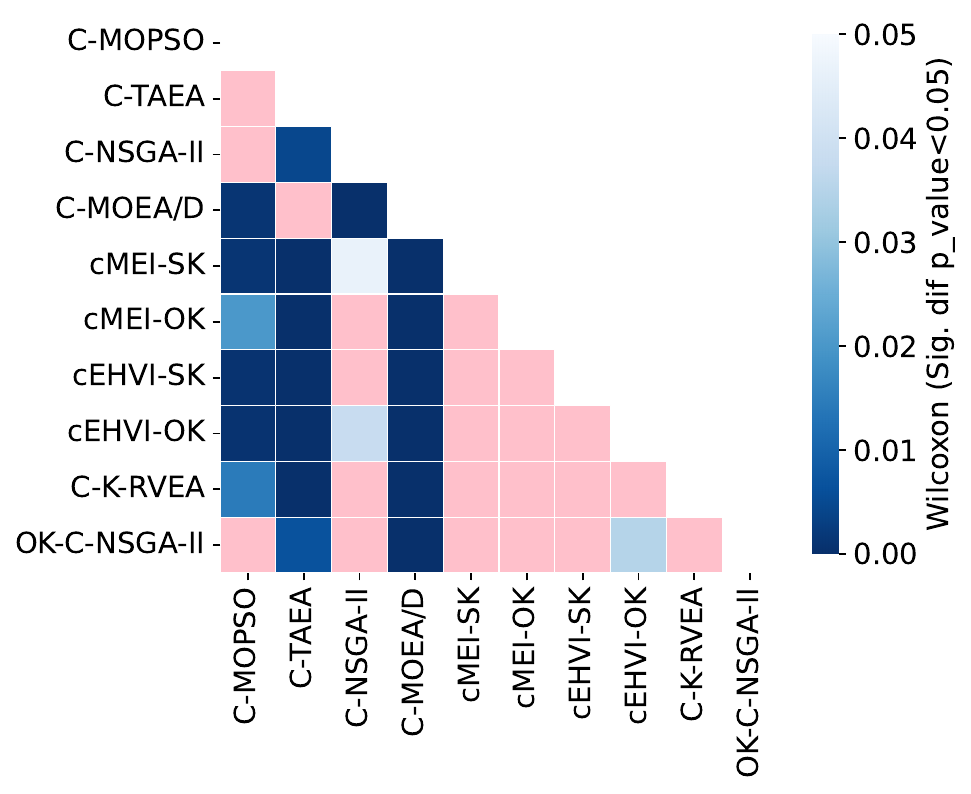}  
\caption{HV}
\label{fig:wilcoxon_hv}
\end{subfigure}
\begin{subfigure}[b]{0.45\textwidth}
\centering
\includegraphics[width=5.5cm]{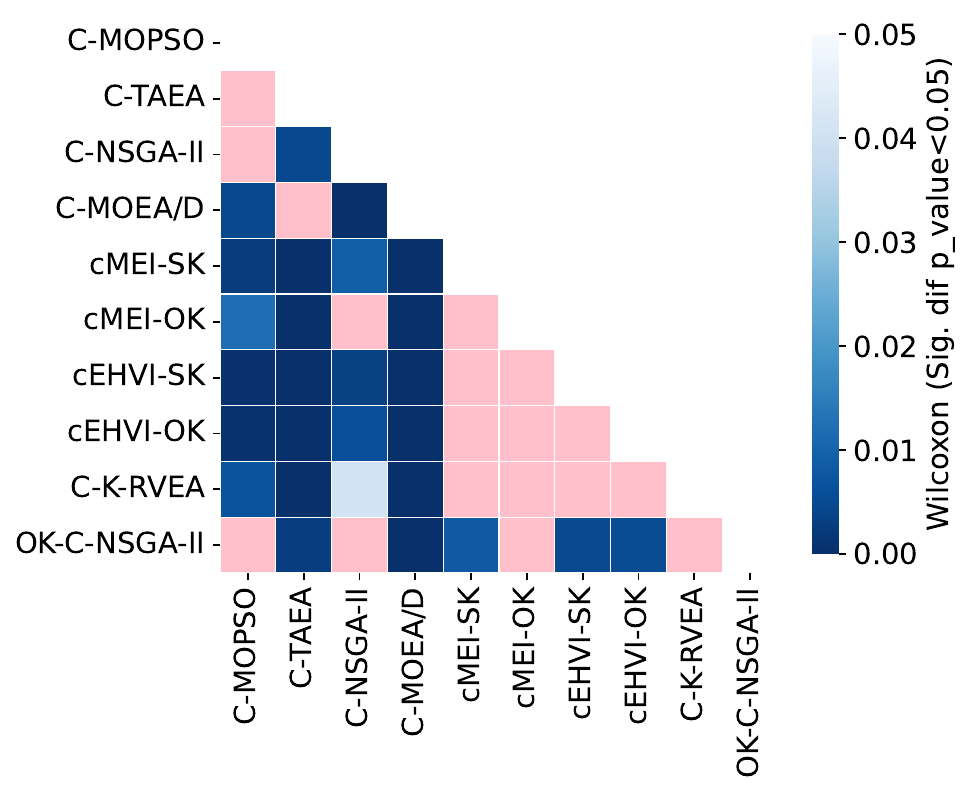}  
\caption{IGD+}
\label{fig:wilcoxon_igd}
\end{subfigure}
\caption{Wilcoxon test results for significant differences between algorithms, using (a) Hypervolume and (b) IGD+ metrics. Pink color indicate no significant differences ($p\_value \ge 5 \%$)}
\label{fig:wilcoxon_test}
\end{figure}

\end{appendices}


\bibliography{references}


\end{document}